%% file: main.tex
\pdfoutput=1

\documentclass[11pt]{article}

\usepackage{booktabs}
\usepackage{multirow}
\usepackage{graphicx}
\usepackage[table,xcdraw]{xcolor}
\usepackage{ulem}

\usepackage{pifont}
\usepackage{soul}
\usepackage[preprint]{acl}

\usepackage{times}
\usepackage{latexsym}

\usepackage[T1]{fontenc}

\usepackage[utf8]{inputenc}

\usepackage{kotex}
\usepackage{amsmath}

\usepackage{microtype}

\usepackage{inconsolata}

\usepackage{graphicx}

\usepackage{xspace}
\newcommand{\sys}{Thunder-LLM\xspace}

\title{\sys: Efficiently Adapting LLMs to Korean \\with Minimal Resources}

\author{
  Jinpyo Kim\thanks{These authors contributed equally to this work.}\thanks{Dept. of Computer Science, Seoul National University} \\
  \texttt{jinpyo@aces.snu.ac.kr} \\\And
  Gyeongje Cho\footnotemark[1]\thanks{Graduate School of Data Science, Seoul National University} \\
  \texttt{gyeongje@aces.snu.ac.kr} \\\And
  Chanwoo Park\footnotemark[1]\footnotemark[2] \\
  \texttt{chanwoo@aces.snu.ac.kr} \\\AND
  Jongwon Park\footnotemark[1]\footnotemark[2] \\
  \texttt{jongwon.park@aces.snu.ac.kr} \\\And
  Jongmin Kim\footnotemark[1]\footnotemark[3] \\
  \texttt{jongmin@aces.snu.ac.kr} \\\AND
  Yeonkyoung So\footnotemark[1]\footnotemark[3] \\
  \texttt{kathy1028@snu.ac.kr} \\\And
  Jaejin Lee\footnotemark[2]\footnotemark[3] \\
  \texttt{jaejin@snu.ac.kr}
  }

\newcommand{\tablecell}[1]{\begin{tabular}[t]{@{}p{\linewidth}@{}}#1\end{tabular}}

\begin{document}
\maketitle

\input{sections/Abstract}
\input{sections/Introduction}

\input{sections/RelatedWork}
\input{sections/Dataset}
\input{sections/Training}
\input{sections/Benchmark}

\input{sections/Evaluation}
\input{sections/Conclusion}

\input{sections/Limitations}
\input{sections/Ethics}
\input{sections/Acknowledgment}


\bibliography{anthology,custom}

\appendix
\input{sections/Appendix-Hyperparameters}
\input{sections/Appendix-Crawling}
\input{sections/Appendix-Post-Training}
\input{sections/Appendix-Benchmark-Eval-Method}

\input{sections/Appendix-ARR-Checklist}

\input{sections/Appendix-Large-Tables}

\end{document}

%% file: sections/Abstract.tex
\begin{abstract}
Since state-of-the-art LLMs often underperform in languages other than English or Chinese, improving the capability of LLMs in new languages has become an essential task. Moreover, LLMs' entire end-to-end training process remains largely unknown to the public due to proprietary reasons, technical complexity, inconsistent documentation, and ethical considerations. The complete picture remains a closely guarded secret within the industry. This paper presents methods to adapt an existing English-based LLM to Korean in a low-budget scenario. We describe the entire end-to-end process: collecting Korean datasets, preprocessing the data, training the model, creating downstream benchmarks, and conducting evaluations. The evaluation results indicate that our method can effectively and cost-efficiently add new language capabilities to existing LLMs. Our new bilingual models, \sys and \sys-Ins, achieve superior Korean performance compared to state-of-the-art models while utilizing minimal data and computational resources. We share our comprehensive experience and make the code publicly available.

\end{abstract}

%% file: sections/Introduction.tex
\section{Introduction}
\label{sec:introduction}
Recent rapid advancements in large language models (LLMs) have made them some of the most powerful tools available today. As a result, the importance of sovereign AI is increasing, with various nations striving to develop their own LLMs that reflect the unique characteristics of their languages and cultures~\cite{glasze2023contested, roberts2023digital}. However, most state-of-the-art LLMs have been developed exclusively by major U.S. or Chinese tech companies and often fail to perform satisfactorily in languages other than English or Chinese~\cite{saura2024digital, saura2024datafeudalism}. For instance, Llama~\cite{grattafiori2024llama}, an open LLM developed by Meta, shows significantly poorer performance in Korean than English. 

While governments, universities, and startups are eager to create LLMs tailored to their specific needs, they frequently lack the necessary hardware resources and technical expertise that large tech companies possess~\cite{izsak-etal-2021-train, gelles2024resource}. Moreover, LLMs' entire end-to-end training process remains largely unknown to the public due to proprietary reasons, technical complexity, inconsistent documentation, and ethical considerations. The complete picture remains a closely guarded secret within the industry. 

Our goal is to train Korean-English bilingual LLMs in a low-budget scenario. We aim to enhance the Korean language capabilities of existing English-based LLMs. However, we have encountered difficulties in finding sufficient resources to improve the language capabilities of these models. Although there have been several attempts to train Korean LLMs, such as those outlined in previous studies~\cite{ko2023technical, choi-etal-2024-optimizing, yoo2024hyperclova, research2024exaone, bak2025kanana}, most of these models and their training methodologies are not publicly available.

Additionally, while several studies have explored adding non-English language capabilities to English-based LLMs~\cite{dou-etal-2024-sailor, cui2023efficient, kiulian2024english, xi2024practice}, we were unable to achieve satisfactory results. These challenges arise from the significant differences in linguistic and cultural characteristics between Korean and other languages.

A significant challenge in training Korean-English LLMs is insufficient data and benchmarks for practical training and evaluation. In contrast to English, which has abundant publicly available data for pre-training LLMs (over several trillion tokens), the amount of high-quality public Korean text is limited to around 30 billion tokens. Moreover, we could not find any public data suitable for post-training Korean LLMs. For evaluation purposes, we need Korean benchmarks encompassing a wide range of domains and task types; however, only a few public benchmarks are available, such as KLUE and KoBEST, which evaluate some specific domains~\cite{son-etal-2025-kmmlu, park2021klue, jang-etal-2022-kobest}.

\begin{figure*}[!t]
	\centering
  	\includegraphics[width=\linewidth]{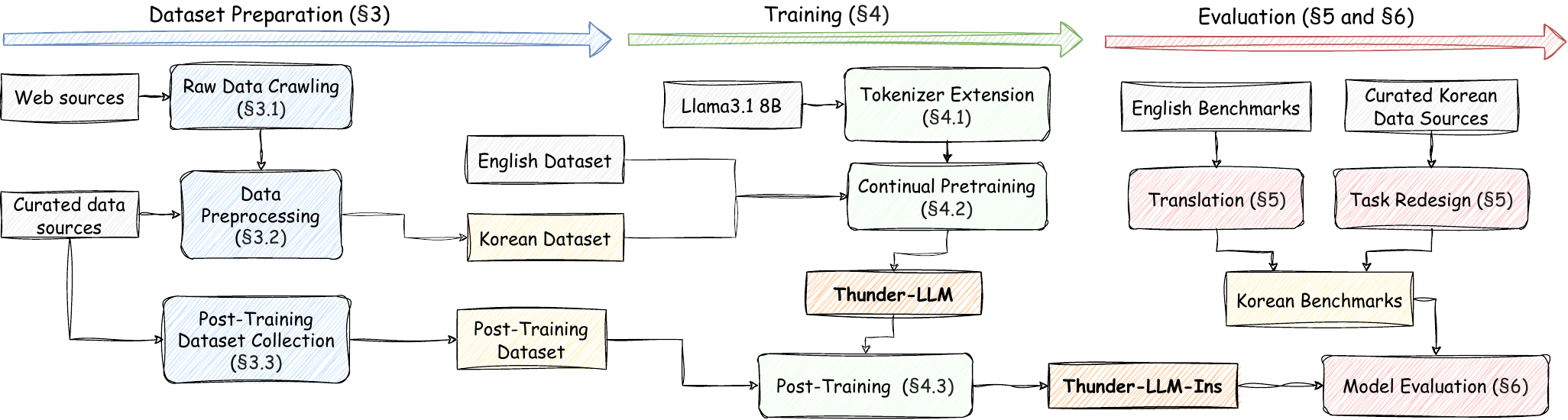}
	\caption{Illustration of the overall process. \label{fig:overview}}
\end{figure*}

To overcome the challenges, we have developed a comprehensive process for training and evaluating bilingual (Korean and English) LLMs, as shown in Figure~\ref{fig:overview}. We start by collecting raw Korean text data and preprocessing it. Using Llama as our baseline English-based LLM, we perform continual pre-training with this collected data to enhance its Korean language capabilities, resulting in the model referred to as \sys\footnote{Complete name of the model is Llama-Thunder-LLM. In this paper, we refer it as Thunder-LLM for simplicity.}. Following this, we conduct additional post-training to further improve the model's performance, producing the final Korean-English bilingual model, identified as \sys-Ins. 

To accelerate the training process, we pinpoint the safe layers within the LLM that can handle low-precision training and implement FP8 training. Lastly, we develop six new Korean benchmarks to evaluate LLM performance in Korean comprehensively.
The evaluation results show that our new models \sys and \sys-Ins achieve the best performance in Korean and comparable performance in English to state-of-the-art models of similar scale, while requiring significantly less training data and computing resources.

This paper discusses our experiences in detail, and we will make the code publicly available. We hope that researchers will use this paper as a foundation to develop new language capabilities for existing language models in low-budget scenarios. Additionally, the proposed method can be applied to low-resource languages other than Korean.

%% file: sections/RelatedWork.tex
\section{Related Work \label{sec:related-work}}

\paragraph{Development of Korean LLMs.}
Several big Korean tech companies have developed their own large language models (LLMs). HyperCLOVA from Naver~\cite{yoo2024hyperclova} and Exaone from LG~\cite{research2024exaone}  are LLMs trained on proprietary data, which allows them to achieve strong performance. However, their reliance on closed datasets hampers reproducibility. Kakao's Kanana~\cite{bak2025kanana}, on the other hand, uses only public data but does not fully disclose its training details, which also limits reproducibility. All of these models are trained from scratch, resulting in high computational costs.

\paragraph{Continual training for bilingual LLMs.}
A cost-effective alternative for training bilingual LLMs is continual training, which builds on an existing LLM to enhance its capabilities. For example, Solar from Upstage~\cite{kim-etal-2024-solar} applies continual training to Llama in order to improve its performance in Korean, although specific details about this process have not been disclosed. Similarly, the bilingual model developed by \citet{gosal2024bilingual} for Arabic, as well as the Sailor model for Southeast Asian languages \cite{dou2024sailor, dou2025sailor2}, use public data and disclosed methods. However, these techniques may not be applicable to Korean due to significant linguistic and cultural differences.


%% file: sections/Dataset.tex
\input{sections/tables/table-english-data-sources}

\section{Preparing Korean Datasets}
This section details our process for collecting and preparing English and Korean datasets to train our language model. For English data, we utilize publicly available, preprocessed datasets such as RedPajama v2~\cite{weber2024redpajama} and DCLM~\cite{li2024datacomp} for general knowledge learning. Additionally, for domain-specific knowledge in English, we source data from Dolma~\cite{soldaini2024dolma}, which includes information from ArXiv, Wikipedia, and similar repositories. A summary of our English data sources can be found in Table~\ref{tab:english-data-sources}. In contrast, due to the lack of sufficient publicly available Korean datasets for training large language models (LLMs), we are compelled to collect raw Korean data from scratch and subsequently preprocess it.

\subsection{Crawling Korean Texts}
We collected a total of 3TB of raw Korean text data from various web sources. A summary of these data sources can be found in Table~\ref{tab:korean-crawl-sources}. The data was gathered from three popular Korean websites: Naver\footnote{\url{https://www.naver.com/}}, Daum\footnote{\url{https://www.daum.net/}}, and Tistory\footnote{\url{https://www.tistory.com/}}. We focused on three types of online content: blogs, online communities (often referred to as cafés in Korea), and news articles. A blog is an informational website consisting of discrete, often informal, posts. An online community (café) is a platform that facilitates discussions on specific topics. News articles, which media companies publish, are frequently distributed through the two major online services, Naver and Daum.

To ensure privacy, we exclude posts with restricted visibility settings,  such as those accessible only to certain members,  from our crawling process. To reduce irrelevant data during the crawling stage, we filter out documents containing specific indicator keywords in their titles. For text extraction (converting HTML to plain text), we have found that fewer than ten HTML structure templates are sufficient to process all the crawled data. Thus, we have developed custom text extraction rules for each HTML template and applied these rules to obtain the raw text data. Further details on the exclusion criteria and text extraction methods can be found in Appendix~\ref{sec:crawling-rules}.


\input{sections/tables/table-korean-crawl-sources}

\subsection{Preprocessing the Dataset}
The raw Korean web dataset contains low-quality documents that are not useful for training LLMs. To enhance the overall quality of the data, we implement a three-stage preprocessing pipeline: (1) rule-based preprocessing, (2) deduplication, and (3) model-based filtering. Each stage is described in detail below.

\paragraph{Rule-based preprocessing.} 
The purpose of this step is to retain only those documents that are natural and meaningful to native Korean speakers. We will only keep documents that meet all of the following criteria, discarding the rest:
\begin{itemize}
    \item  A document must contain between 10 and 10,000,000 words. This requirement helps filter out documents that are either too short to provide meaningful training signals or excessively long and noisy. 
\item The average word length in a document must range from 2 to 10 characters. This criterion eliminates documents that are unlikely to consist of natural Korean words. 

\item At least 80\% of the words in a document must be in Korean characters. This ensures that the training corpus focuses primarily on Korean-dominant content. 

\item The most frequent 5-gram in a document should not account for more than 15\% of all 5-grams present. This helps to eliminate spam-like or overly repetitive documents.
\end{itemize}

These filtering conditions are simpler than those used in previous studies~\cite{soldaini2024dolma, weber2024redpajama, rae2021scaling} because our web data crawling process includes HTML tag removal and a boilerplate detection process.

We then apply a regex-based rule that removes sentences that lack ending punctuation marks, such as '.', '?', and '!'. This step helps eliminate boilerplate content, like image captions and copyright notices, which often appear in news articles. By applying all these filtering rules, we filter out 45\% of the raw crawling data.

\paragraph{Deduplication.} 
We perform deduplication based on document similarity to eliminate redundant content and shared templates across websites. This deduplication is essential in preprocessing web-crawled data, as certain content can dominate the dataset, negatively impacting both its diversity and quality~\cite{son2025fed}. We use a GPU-based deduplication technique developed by \citet{son2025fed}, which efficiently handles large-scale datasets by calculating document hashes and clustering similar content. As a result, we were able to remove 10.7\% of our data that was duplicated.

\paragraph{Model-based filtering.}
\label{sec:model-based-filtering}
We implement model-based filtering to select documents that provide the richest and most coherent contexts. To achieve this, we train a 5-gram KenLM language model~\cite{heafield2011kenlm}  on a dump of the Korean Wikipedia~\cite{wikidump}. We then filter out documents from the collected web data that exhibit high perplexity, as this indicates a low likelihood under the language model and suggests poor linguistic fluency and naturalness. The threshold for filtering out high perplexity documents is set based on our computational budget, specifically the number of tokens that can be processed during model training. This ensures that the filtered dataset retains as much useful data as possible while minimizing perplexity.

\subsection{Datasets for Post-training \label{sec:dataset-post-training}}
Similar to the pre-training phase, there is a lack of publicly available Korean datasets for post-training. We adopt two approaches to collect datasets for Supervised Fine-Tuning (SFT)~\cite{ouyang2022training} and Direct Preference Optimization (DPO)~\cite{rafailov2023direct}. One approach involves using the training datasets from each language model benchmark, while the other uses synthetic data~\cite{qwen25technicalreport, research2024exaone}. In this section, we briefly explain the methods used to construct the post-training datasets, and additional details can be found in Appendix~\ref{sec:post-training-details}.

\paragraph{Training set of benchmarks.}
We collect the training datasets from public Korean and English language model benchmarks and convert them into a question-and-answer format for SFT. For DPO, we gather training datasets from multiple-choice language model benchmarks and create preference datasets. In these datasets, the correct answer to each question is treated as the chosen response, while the incorrect answers are considered the rejected responses.

\paragraph{Synthetic datasets.}
\label{para:synthetic-datasets}
We gather high-quality questions from online sources and generate responses using a language model. The full list of online sources can be found in Appendix~\ref{sec:post-training-details}. Approaches that exploits language models to produce responses are common for creating post-training datasets~\cite {grattafiori2024llama, research2024exaone}. 

We utilize the Llama3.3-70B-Instruct \cite{grattafiori2024llama}, EXAONE3.5-32B-Instruct \cite{research2024exaone}, and QWen2.5-32B and 72B model families \cite{qwen25technicalreport} to generate responses for each question. Initially, we filter out low-quality responses, including those that are excessively long or not written in the target languages (Korean and English). Next, we review the responses and incorporate the correct ones into our SFT and DPO datasets. Incorrect responses are categorized as rejected responses for DPO.

%% file: sections/tables/table-english-data-sources.tex

\begin{table}[htbp]
\centering
\setlength{\tabcolsep}{4pt}
\renewcommand{\arraystretch}{1}
\resizebox{0.8\columnwidth}{!}{%
\begin{tabular}{@{}ccrr@{}}
\toprule
\textbf{Source} & \textbf{Type} & \textbf{Size} & \textbf{\# Tokens} \\ \midrule \midrule
Redpajama v2    & Web           & 1.2 TB               & 1.0 T                 \\ \midrule
DCLM            & Web           & 5.9 TB               & 3.8 T               \\ \midrule
Dolma           & Curated       & 321.3 GB             & 169.3 B             \\ \bottomrule
\end{tabular}%
}
\caption{Sources of the English datasets.
\label{tab:english-data-sources}}
\end{table}

%% file: sections/tables/table-korean-crawl-sources.tex
\begin{table}[htbp]
\centering
\setlength{\tabcolsep}{3pt}
\renewcommand{\arraystretch}{0.9}
\resizebox{0.9\columnwidth}{!}{%
\begin{tabular}{@{}ccrl@{}}
\toprule
\textbf{Source} & \textbf{Type} & \textbf{Size} & \begin{tabular}[c]{@{}c@{}}
\textbf{Subcategory} 
\end{tabular}
\\ \midrule  \midrule
Naver           & Web           & 2.7 TB         & Blog, Cafe, News and Q\&A                                                                            \\ \midrule
Tistory         & Web           & 16.7 GB        & Blog                                                                                                 \\ \midrule
Daum            & Web           & 624.8 GB       & Cafe and News                                                                                        \\ \midrule
AiHub           & Curated       & 17.4 GB        & \begin{tabular}[c]{@{}l@{}}Scientific Articles, Wiki,\\ Book summary and Synthetic chat\end{tabular} \\ \midrule
KISTI           & Curated       & 27.9 GB        & \begin{tabular}[c]{@{}c@{}}Scientific Articles Summary \\ and Scientific Q\&A\end{tabular}           \\ \bottomrule
\end{tabular}%
}
\vspace{-0.5\baselineskip}
\caption{Crawling sources for Korean texts. 
\label{tab:korean-crawl-sources}}
\end{table}

%% file: sections/Training.tex
\section{Training Methods}
This section outlines our methods for training our Korean-English bilingual model. We use Llama3.1-8B~\cite{grattafiori2024llama} as our baseline model. To enhance its performance, we replace and extend the tokenizer by adding additional Korean tokens. We then continue training the model using our Korean and English datasets. Following this, we conduct a brief additional training session to address any weaknesses in specific domains. Lastly, we perform post-training on the model.

\subsection{Extension of the Tokenizer 
\label{sec:model-arch}}

\begin{figure}[htbp]
	\centering
	\includegraphics[width=0.92\columnwidth]{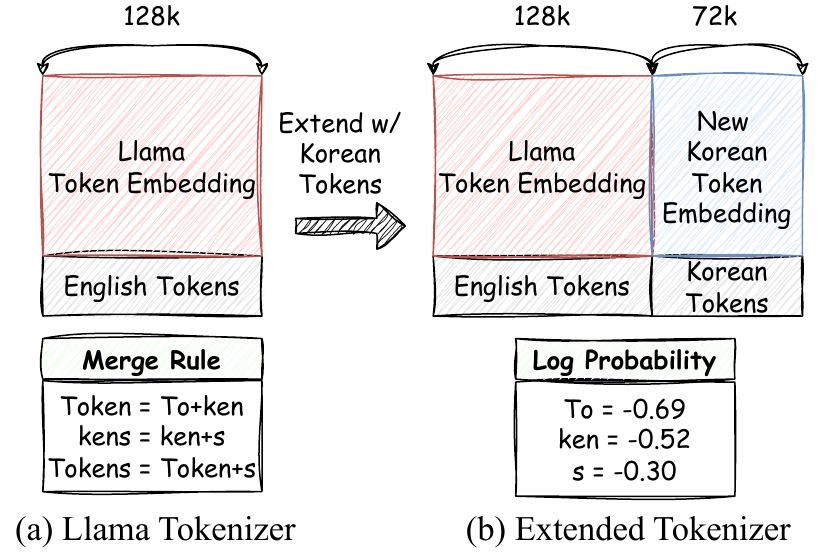}
	\caption{Extending the original Llama tokenizer with new Korean tokens. \label{fig:tokenizer}}
\end{figure}

We extend the original Llama 3.1 tokenizer with new Korean tokens to lower inference costs while maintaining accuracy in non-Korean tasks, as illustrated in Figure~\ref{fig:tokenizer}. We develop a tokenizer extension method\footnote{The paper on this topic is under review. We will add a citation after the paper is published.}, which introduces a Korean-optimized pre-tokenization strategy based on branching entropy for vocabulary construction. We first create a Korean-specific tokenizer with a vocabulary size of 72,000 tokens, then add these tokens to the original Llama tokenizer, resulting in a combined vocabulary of 200,000 tokens. The Llama's original tokens remain the same. The embedding vector for each new Korean token is initialized by averaging the embedding vectors of the sub-tokens generated by tokenizing the new token with the original Llama tokenizer~\cite{minixhofer2021wechsel}.

We switch the tokenization algorithm from Byte-Pair Encoding (BPE) to the Unigram tokenizer to tokenize Korean text effectively. It selects the tokenization path that maximizes the total log probability of tokens in a sentence by using the log probabilities assigned to each token. We calculate the probabilities based on the tokenization results from the original Llama tokenizer to maintain performance in English and other languages besides Korean. This approach ensures consistency with the original tokenizer for non-Korean inputs.





\subsection{Continual Pre-training 
\label{sec:continual-pretraining}}
Continual pre-training significantly enhances the performance of English-based LLMs on non-English tasks~\cite{choi2024optimizing, gosal2024bilingual}. Training foundation models from scratch is often computationally expensive; therefore, it is common to build upon open-source models like Llama, which offer publicly available parameters. In this study, we adopt this approach to improve Llama's capabilities in Korean while maintaining its effectiveness on English tasks.

We begin by conducting a hyperparameter search focusing on batch size and learning rate. To analyze performance trends, we train the models on only 1B tokens, which represents roughly 1\% of the full training set. After this preliminary phase, we select the best-performing configuration based on the evaluation results of the downstream task. Then, we proceed with full training, using a total of 102B. We maintain an approximate 1:1 ratio of English to Korean texts. This ratio is chosen empirically to enhance Korean performance quickly while minimizing any adverse effects on English performance. The training data is evenly split between academic texts and web sources. Additional details and hyperparameters for the continual pre-training phase are summarized in \autoref{tab:hyperparameters} of Appendix~\ref{sec:hyperparameters}.

\subsection{Post-Training 
\label{sec:post-training}}
After task-specific training, we perform post-training on the model using Supervised Fine-Tuning (SFT) and Direct Preference Optimization (DPO) methods to further enhance downstream task performance.
We follow the method developed by \citet{wei2022finetunedlanguagemodelszeroshot} for SFT, and method by \citet{meng2024simposimplepreferenceoptimization} for DPO.
Detailed statistics and hyperparameters for SFT and DPO are summarized in \autoref{tab:hyperparameters} of Appendix~\ref{sec:hyperparameters}.
To ensure balanced learning across different domains, we collect our post-training dataset from various sources. We sample an equal amount of data from each source. For those sources that do not have enough data, we duplicate samples to reach the desired number of training examples, maintaining balance across all domains.

\subsection{Training Platform \label{sec:training-platform}}


\paragraph{Codebase.}
We have developed an in-house framework based on PyTorch~\cite{paszke2019pytorch}, which encompasses the entire process, including dataset preparation, model training, and evaluation. To parallelize the training process, we utilize the DeepSpeed~\cite{rajbhandari2020zero} framework. The codebase will be made publicly available.

\input{sections/tables/table-benchmarks}

\input{sections/tables/table-fp8-stability}
\paragraph{Training in FP8.}
Modern GPUs come equipped with specialized hardware, such as 4th-generation Tensor Cores~\cite{nvidiah100}, which are optimized for FP8 matrix multiplications. The FP8 data type can deliver up to twice the throughput compared to BF16 or FP16. Since a significant portion of LLM training time is devoted to matrix multiplications, exploiting FP8 precision for compatible layers can considerably reduce overall training time.

However, applying FP8 precision to all matrix multiplications in LLMs can lead to training instability, often resulting in loss explosion. While there have been previous studies aimed at addressing this instability~\cite{fishman2024scaling, peng2023fp8}, they often require modifications to the model architecture or the training process. In contrast, our approach is more straightforward: we identify the specific layers that cause instability and apply FP8 precision only to those layers that are confirmed to be stable.

We categorize matrix multiplications in Transformer models into three types based on the layer type: linear layers within Transformer blocks, matrix multiplications in attention mechanisms, and the language model head. We then evaluate the stability of FP8 training for each type by switching each layer between FP8 and BF16. To conduct this assessment, we train a small Llama-like Transformer model with 360M parameters until the loss either converges or becomes unstable. The results, presented in \autoref{tab:fp8-stability}, indicate that using FP8 for matrix multiplications related to attention mechanisms can lead to instability. However, FP8 can be safely used for the other layers without compromising training stability.


Using FP8 precision in such a way can achieve a $\times1.4$ increase in end-to-end training speed compared to traditional FP16 or BF16 training, without sacrificing model accuracy. We employ the Transformer Engine framework~\cite{nvidia_transformer_engine} for FP8 training of our models.

%% file: sections/tables/table-benchmarks.tex
\begin{table*}[t]
\centering

\setlength{\tabcolsep}{3pt}
\renewcommand{\arraystretch}{0.9}
\resizebox{0.9\textwidth}{!}{%
\begin{tabular}{@{}l|l|l||l|l|l@{}}
\toprule
\multicolumn{3}{c||}{\textbf{Pre-training Benchmarks}}                                                                                      & \multicolumn{3}{c}{\textbf{Post-training Benchmarks}}                                                                                                           \\
\multicolumn{1}{c|}{\textbf{Domain}}                                                                              & \multicolumn{1}{c|}{\textbf{English}} & \multicolumn{1}{c||}{\textbf{Korean}}         & \multicolumn{1}{c|}{\textbf{Domain}}                                                                             & \multicolumn{1}{c|}{\textbf{English}}         & \multicolumn{1}{c}{\textbf{Korean}}                     \\ \hline \hline
\multirow{3}{*}{\begin{tabular}[c]{@{}l@{}}General language\\ understanding\end{tabular}}     & HellaSwag        & KoBEST-HellaSwag        & \multirow{3}{*}{Math problem solving}                                                       & \multirow{3}{*}{GSM8K}     & \multirow{3}{*}{Ko-GSM8K$^\ast$}     \\
                                                                                              & WinoGrande       & Ko-WinoGrande$^\ast$    &                                                                                             &                            &                                      \\
                                                                                              & -                & Ko-LAMBADA$^\dagger$    &                                                                                             &                            &                                      \\ \hline
\multirow{2}{*}{\begin{tabular}[c]{@{}l@{}}Scientific knowledge\\ understanding\end{tabular}} & ARC-Easy         & Ko-ARC-Easy$^\ast$      & \multirow{2}{*}{\begin{tabular}[c]{@{}l@{}}Instruction following\\ capability\end{tabular}} & \multirow{2}{*}{IFEval}    & \multirow{2}{*}{Ko-IFEval$^\ast$}    \\
                                                                                              & ARC-Challenge    & Ko-ARC-Challenge$^\ast$ &                                                                                             &                            &                                      \\ \hline
\begin{tabular}[c]{@{}l@{}}
Academic knowledge\\ 
understanding\end{tabular}                    & MMLU             & KMMLU                   & \multirow{2}{*}{Coding capability}                                                          & \multirow{2}{*}{HumanEval} & \multirow{2}{*}{KR-HumanEval} \\ \cline{1-3}
OpenbookQA                                                                                    & OpenbookQA       & -                       &                                                                                             &                            &                                      \\ \bottomrule
\end{tabular}%
}
\caption{Downstream benchmarks used in the model evaluation. $^\ast$ indicates the benchmarks we built by translation.  $^\dagger$ indicates the benchmarks we built from scratch. \label{tab:benchmarks}}
\end{table*}

%% file: sections/tables/table-fp8-stability.tex
\begin{table}[htbp]
\centering
\setlength{\tabcolsep}{3pt}
\renewcommand{\arraystretch}{0.9}
\resizebox{0.7\columnwidth}{!}{%
\begin{tabular}{@{}cccc@{}}
\toprule
\textbf{Attention} & \textbf{Linear} & \textbf{LM Head} & \textbf{Stable Training} \\ \midrule \midrule
BF16               & BF16            & BF16            & Yes                   \\
BF16               & BF16            & FP8             & Yes                   \\
BF16               & FP8             & BF16            & Yes                   \\
BF16               & FP8             & FP8             & Yes                  \\ \midrule
FP8                & BF16            & BF16            &  No                        \\
FP8                & BF16            & FP8             &  No                     \\
FP8                & FP8             & BF16            &  No                        \\
FP8                & FP8             & FP8             &  No                        \\ \bottomrule
\end{tabular}%
}
\vspace{-0.5\baselineskip}
\caption{Results of training stability test for matrix multiplications in different layer components. 
\label{tab:fp8-stability}}
\end{table}

%% file: sections/Benchmark.tex
\input{sections/tables/table-korean-result}
\section{Downstream Benchmarks}
We use six Korean downstream benchmarks~\citep{jang-etal-2022-kobest, son2024kmmlu} and six English benchmarks~\citep{zellers2019hellaswag, sakaguchi2021winogrande, clark2018think, hendrycks2021mmlu, mihaylov-etal-2018-suit} to assess the performance of pre-trained LLMs. For models that undergo post-training, we incorporate an additional three downstream benchmarks for both Korean and English~\citep{cobbe2021gsm8k, zhou2023instruction, chen2021humaneval, kang2024krhumaneval}. A complete list of the benchmarks can be found in \autoref{tab:benchmarks}. Detailed information on the evaluation methods for each benchmark is provided in Appendix~\ref{sec:benchmark-eval-method}.

Unlike English language models, which have numerous well-established benchmarks for evaluation, there is a lack of publicly available datasets for assessing Korean language models. Although there are specialized evaluation datasets for Korean, such as KLUE~\cite{park2021klue} and KoBEST~\cite{jang-etal-2022-kobest}, a comprehensive assessment that spans a wide range of fields and task types is still insufficient. This gap makes the evaluation of Korean language models inconsistent and challenging.

To address the shortage of Korean benchmarks, we have created a total of six new downstream benchmarks. Out of these, five (Ko-ARC-E, Ko-ARC-C, Ko-GSM8K, Ko-WinoGrande, and Ko-IFEval) are translations of existing English benchmarks~\cite{clark2018think, ham-etal-2020-kornli, cobbe2021training, sakaguchi2021winogrande, zhou2023instruction}. 
We begin by using DeepL\footnote{\url{https://www.deepl.com/en/products/api}} for machine translation and then proceed with human revision and localization. This revision process is carried out by domain experts who correct any inaccurate translations and ensure consistency in writing style and expressions. The translated benchmarks are also localized to reflect Korean cultural and linguistic characteristics, which include adapting personal names, place names, and measurement units, as well as revising foreign cultural references and concepts that may be unfamiliar to Korean audiences~\citep{choi-etal-2024-optimizing}.

In addition, we are introducing a completely new benchmark for Korean, called Ko-LAMBADA. The original LAMBADA benchmark~\cite{paperno-etal-2016-lambada} based on English literary texts, presents significant challenges when translating into Korean, even with extensive revisions. Additionally, there are notable linguistic differences between English and Korean. The LAMBADA benchmark evaluates a language model's ability to predict the last word of a sentence, which is typically a noun or a person's name in English. However, in Korean, sentences usually end with verbs, making the prediction of the last word less effective for assessing contextual understanding. Therefore, we have redesigned the task for Korean to focus on predicting important words, such as nouns that appear in the middle of the sentence.

All Korean benchmarks we build are cross-checked by additional independent reviewers who did not participate in the initial revision and localization process to identify and correct any errors in the datasets.

%% file: sections/tables/table-korean-result.tex
\begin{table*}[t]
\centering
\setlength{\tabcolsep}{3pt}
\renewcommand{\arraystretch}{0.9}
\resizebox{\textwidth}{!}{%
\begin{tabular}{@{}c|ccccccccc|c@{}}
\toprule
\textbf{Model}             & \textbf{\begin{tabular}[c]{@{}c@{}}KoBEST-\\ HellaSwag\end{tabular}} & \textbf{\begin{tabular}[c]{@{}c@{}}Ko-\\ WinoGrande\end{tabular}} & \textbf{\begin{tabular}[c]{@{}c@{}}Ko-\\ LAMBADA\end{tabular}} & \textbf{\begin{tabular}[c]{@{}c@{}}Ko-\\ ARC-E\end{tabular}} & \textbf{\begin{tabular}[c]{@{}c@{}}Ko-\\ ARC-C\end{tabular}} & \textbf{KMMLU} & \textbf{\begin{tabular}[c]{@{}c@{}}Ko-\\ GSM8K\end{tabular}} & \textbf{\begin{tabular}[c]{@{}c@{}}Ko-\\ IFEval\end{tabular}} & \textbf{\begin{tabular}[c]{@{}c@{}}KR-\\ HumanEval\end{tabular}} & \textbf{Avg.} \\ \midrule
Llama-3.1-8B               & 58.2                                                                 & 60.6                                                              & 84.3                                                           & 63.3                                                         & 44.6                                                         & 40.5           & 34.6                                                         & 30.7                                                          & 21.9                                                             & 48.7          \\
\rowcolor[HTML]{E3FBE3} 
\sys                       & 59.8                                                                 & 61.3                                                              & \textbf{87.9}                                                  & 70.1                                                         & 49.1                                                         & 41.3           & 32.5                                                         & 30.5                                                          & 42.1                                                             & 52.7          \\
\rowcolor[HTML]{ECF4FF} 
\sys-Ins                   & \textbf{72.4}                                                        & \textbf{74.3}                                                     & \uline{86.8}                                                   & \uline{76.1}                                                 & \textbf{62.4}                                                & 47.6           & 57.3                                                         & 51.5                                                          & \uline{56.7}                                                    & \textbf{65.0} \\ \midrule
Llama-3.1-8B-Instruct      & 55.8                                                                 & 60.2                                                              & 83.8                                                           & 64.4                                                         & 45.7                                                         & 41.1           & 53.1                                                         & 43.4                                                          & 42.1                                                             & 54.4          \\
EXAONE-3.5-7.8B-Instruct   & 60.0                                                                 & \uline{65.3}                                                      & 85.7                                                           & \textbf{76.7}                                                & \uline{57.0}                                                 & 45.1           & 56.7                                                         & \textbf{67.9}                                                 & \uline{\textbf{61.0}}                                           & \uline{63.9}  \\
DNA-1.0-8B-Instruct        & \uline{61.2}                                                         & 63.4                                                              & 80.0                                                           & 72.7                                                         & 55.7                                                         & \textbf{53.1}  & 58.5                                                         & 44.4                                                          & 28.1                                                             & 57.5          \\
Qwen2.5-7B-Instruct        & 58.2                                                                 & 63.7                                                              & 81.7                                                           & 69.4                                                         & 54.5                                                         & \uline{49.6}     & \textbf{67.3}                                                & \uline{60.5}                                                 & 28.1                                                             & 59.2          \\
Ministral-8B-Instruct-2410 & 59.6                                                                 & 62.3                                                              & 84.9                                                           & 67.7                                                         & 50.3                                                         & 40.1           & \uline{59.5}                                                   & 37.0                                                          & 14.0                                                             & 52.8          \\
Gemma-7b-it                & 56.2                                                                 & 62.2                                                              & 81.4                                                           & 67.1                                                         & 48.2                                                         & 34.3           & 36.2                                                         & 30.3                                                          & 29.9                                                             & 49.5          \\ \bottomrule
\end{tabular}

}
\caption{Korean benchmark performance of the models.
The bold-faced text represents the highest
performance, and the underlined text represents the second highest performance of each benchmark. 
\label{tab:korean-result}}
\end{table*}

%% file: sections/Evaluation.tex
\section{Evalution}
\label{sec:evaluation}
This section evaluates the performance of the \sys and \sys-Ins models, which were trained using the proposed datasets and training methods. After each training stage, the benchmark performance of the models in both Korean and English is presented in \autoref{tab:korean-result} and \autoref{tab:english-result}, respectively. We compare the performance of our models with state-of-the-art models of a similar scale that support Korean functionalities. Notable models include EXAONE-3.5 (8B)~\cite{research2024exaone} and DNA 1.0 (8B Instruct)~\cite{lee2025dna10technicalreport}, which are Korean-English bilingual models. Additionally, we consider multilingual LLMs, such as Qwen2.5 (7B Instruct)~\cite{qwen25technicalreport}, Mistral-8B Instruct~\cite{jiang2023mistral7b}, and Gemma-7B IT~\cite{team2024gemma}, which also support Korean.

Compared to the state-of-the-art 8B-scale models that use several trillion tokens for training, we only use around 100 billion tokens for continual pre-training and a few million tokens for post-training.

\subsection{Pre-Trained Model Performance}

The model, referred to as \sys, is derived from the tokenizer extension and the continual pre-training of the baseline Llama3.1 8B. Evaluation results indicate that continual pre-training improves performance on Korean benchmarks by an average of 4\%. This demonstrates the effectiveness of our data collection and training strategy. In contrast, the English benchmark scores after continual pre-training show only minimal variation compared to the baseline Llama3.1 8B model. 

Our approach significantly improves Korean capabilities while maintaining the model's original performance. Among the English benchmarks, ARC-Challenge, MMLU, and GSM8K show relatively significant declines in performance. This is due to the limited general knowledge present in our English training corpus, and we plan to address this issue through post-training.

\subsection{Post-Trained Model Performance}
\input{sections/tables/table-english-result}

\sys-Ins is the model we use for post-training on \sys. As anticipated, the benchmark scores after post-training showed significant improvements in both Korean and English. Overall, \sys-Ins outperforms other models in Korean and ranks second in English. It excels in several benchmarks, particularly in general language understanding.

We observe a significant performance improvement in the benchmarks that include their training set in our post-training dataset. Additionally, there is a marked increase in the benchmark scores for those that do not have their training datasets, such as IFEval and Ko-IFEval, included in the post-training dataset. This evidence confirms the effectiveness of our post-training dataset collection and training methods.

\begin{table}[htbp]
\centering
\setlength{\tabcolsep}{3pt}
\renewcommand{\arraystretch}{0.9}
\resizebox{0.7\columnwidth}{!}{%
\begin{tabular}{@{}cc|cc@{}}
\toprule
\textbf{Model}          & \textbf{\# Params}  & \textbf{Precision} & \textbf{Tokens/sec} \\ \midrule
\multirow{2}{*}{\sys} & \multirow{2}{*}{8B} & BF16               & 0.23M               \\ \cmidrule(l){3-4} 
                        &                     & FP8                & 0.33M               \\ \bottomrule
\end{tabular}%
}
\vspace{-0.5\baselineskip}
\caption{Training speed of \sys models using FP8 or BF16 precision for matrix multiplications. We use the same training configuration with the continual pre-training stage, as described in Table~\ref{tab:hyperparameters} of Appendix~\ref{sec:hyperparameters}.
\label{tab:fp8-result}}
\end{table}

\subsection{Training Speed}

\paragraph{Impact of FP8 training.}
We evaluate the training speed of \sys models, which utilize the FP8 training method described in Section~\ref{sec:training-platform}. Consequently, training \sys models in FP8 results in a speedup of 1.43  compared to traditional BF16 training. Additionally, we observe no drop in model accuracy when training them in FP8.

\input{sections/tables/table-inference-speed}

\subsection{Inference Speed}
\paragraph{Impact of tokenizer extension.}
We evaluate the effect of the tokenizer extension discussed in Section~\ref{sec:model-arch} on inference speed. Our assessment focuses on the time needed to complete the HellaSwag (English) and KoBEST-HellaSwag (Korean) benchmarks using Llama-3.1-8B and \sys, as decribed in \autoref{tab:inference-speed}.

For the Korean benchmark, the number of tokens needed for evaluation is nearly halved, resulting in an 18.0\% reduction in total inference time. We conclude that improving the LLM's tokenizer using the Korean vocabulary construction method outlined in Section~\ref{sec:model-arch} significantly speeds up LLM inference in Korean.

The number of tokens remains roughly consistent for the English benchmark, but the inference time increases by 7.2\%. This rise is due to the larger language model (LM) head that results from the increased vocabulary size, leading to higher computational costs. However, this slowdown becomes less significant when the model is scaled up in parameters, as the computational cost of the LM head becomes smaller relative to that of the overall Transformer architecture.



%% file: sections/tables/table-english-result.tex
\begin{table*}[t]
\centering
\setlength{\tabcolsep}{3pt}
\renewcommand{\arraystretch}{0.9}
\resizebox{\textwidth}{!}{%
\begin{tabular}{@{}c|ccccccccc|c@{}}
\toprule
\textbf{Model}             & \textbf{HellaSwag} & \textbf{WinoGrande} & \textbf{OBQA} & \textbf{ARC-E} & \textbf{ARC-C} & \textbf{MMLU} & \textbf{GSM8K} & \textbf{IFEval} & \textbf{HumanEval} & \textbf{Avg.} \\ \midrule
Llama-3.1-8B               & 79.0               & 77.0                & 44.8          & 91.5           & 79.5           & 65.3          & 57.2           & 18.7            & 34.8               & 60.9          \\
\rowcolor[HTML]{E3FBE3} 
\sys                       & 79.0               & 77.9                & 44.8          & 90.7           & 77.6           & 62.9          & 53.9           & 32.6            & \textbf{69.5}      & 65.4          \\
\rowcolor[HTML]{ECF4FF} 
\sys-Ins                   & \textbf{89.3}      & \textbf{89.4}       & \textbf{64.0} & 91.3           & 80.3           & 63.1          & 76.5           & 59.1            & 59.1               & \uline{74.7}  \\ \midrule
Llama-3.1-8B-Instruct      & 79.2               & 78.1                & 43.0          & 93.3           & 83.2           & \uline{68.0}  & 77.2           & 61.4            & 57.3               & 71.2          \\
EXAONE-3.5-7.8B-Instruct   & 77.9               & 74.4                & 46.4          & \uline{95.4}   & \uline{85.5}   & 65.2          & 73.7           & \textbf{78.4}   & \uline{66.5}       & 73.7          \\
DNA-1.0-8B-Instruct        & 79.9               & 77.0                & 45.8          & 93.5           & 83.6           & 66.6          & \uline{81.7}   & 55.0            & 34.8               & 68.7          \\
Qwen2.5-7B-Instruct        & 80.4               & 74.6                & \uline{48.6}  & \textbf{96.7}  & \textbf{90.3}  & \textbf{74.2} & \textbf{83.1}  & \uline{74.8}    & 64.6               & \textbf{76.4} \\
Ministral-8B-Instruct-2410 & 79.1               & \uline{79.1}        & 46.8          & 93.6           & 83.6           & 64.8          & 77.3           & 37.5            & 57.9               & 68.9          \\
Gemma-7b-it                & \uline{80.6}       & 77.0                & 44.0          & 92.6           & 79.6           & 62.7          & 55.0           & 36.1            & 33.5               & 62.3          \\ \bottomrule
\end{tabular}
}
\caption{English benchmark performance of the models.
The bold-faced text represents the highest
performance, and the underlined text represents the second highest performance of each benchmark. \label{tab:english-result}}
\end{table*}

%% file: sections/tables/table-inference-speed.tex
\begin{table}[htbp]
\centering
\setlength{\tabcolsep}{3pt}
\renewcommand{\arraystretch}{0.9}
\resizebox{0.85\columnwidth}{!}{%
\begin{tabular}{@{}c|cc|cc@{}}
\toprule
\multirow{2}{*}{\textbf{Tokenizer}} & \multicolumn{2}{c|}{\textbf{HellaSwag}} & \multicolumn{2}{c}{\textbf{KoBEST-HellaSwag}} \\
                                    & \# Tokens           & Time (s)           & \# Tokens              & Time (s)              \\ \midrule
Llama                               & 3.20 M              & 1,713.8             & 204.0 K                & 147.3                 \\
Extended                           & 3.19 M              & 1,838.4             & 108.8 K                & 120.9                 \\ \bottomrule
\end{tabular}%
}
\vspace{-0.5\baselineskip}
\caption{Comparing inference speed between the original Llama and Korean-extended tokenizers when used in Llama-3.1-8B and \sys, respectively.  \label{tab:inference-speed}}
\end{table}


%% file: sections/Conclusion.tex
\section{Conclusions}
\label{sec:conclusion}
This paper presents a cost-effective, end-to-end process for training LLMs to improve the Korean capabilities of English-based multilingual models. We begin by collecting and preprocessing Korean text data. Using the English-based Llama foundational model, we enhance its tokenizer with a Korean vocabulary construction method. Next, we continually pre-train and post-train the model in FP8 format using various datasets to improve its performance in Korean. We also develop six new Korean downstream benchmarks to address the lack of benchmarks for evaluating Korean LLMs. 
Our models, \sys and \sys-Ins, demonstrate the best performance in Korean and achieve comparable results in English when measured against state-of-the-art models. Notably, they do this while requiring significantly less data and computational resources. We intend to publicly release the code and model parameters to provide a valuable reference for other researchers and developers.

%% file: sections/Limitations.tex
\section*{Limitations 
\label{sec:limitations}}
We do not aim to build a state-of-the-art LLM. Instead, our goal is to demonstrate an effective and reproducible training methodology for enhancing Korean language capabilities within existing English-based LLMs. We anticipate that the model's performance will improve further with additional resources, such as increased computational power and human effort for data refinement. 

Although we use both Korean and English data during continual pre-training, we have observed a slight decline in English performance. This is likely due to the imbalance in data quality and quantity, as our primary focus has been on collecting Korean datasets. We believe this issue can be addressed by incorporating more English datasets to enhance general knowledge in future training. 

While we plan to release all source code and tools used for data collection and training, we cannot share the actual collected datasets due to copyright and privacy concerns. 

The effectiveness of our method has only been validated for the Korean language. Further experiments are necessary to assess its applicability to other low-resource or typologically distant languages. 

Due to computational constraints, all experiments were conducted with models containing up to 8 billion parameters. We will reserve the evaluation of larger-scale language models for future work.

%% file: sections/Ethics.tex
\section*{Ethics Statement \label{sec:ethics}}

We construct the Ko-LAMBADA benchmark using only public domain texts, primarily classical literary works whose copyrights have expired. Since the content is fictional, the dataset does not contain personal information or real-world references that may raise ethical concerns. Therefore, there are no copyright issues.

We collect web data only from sites that do not implement technical restrictions against crawling. We exclude any content that is not publicly accessible and ensure that our crawling process does not impose excessive load on the target servers. Hence, our data collection does not have legal concerns.

We collect data only for research purposes and will not distribute the collected dataset. The released models are intended solely for research use, and we take necessary steps to respect and protect the copyright of original content owners.

%% file: sections/Acknowledgment.tex
\section*{Acknowledgments}
This work was partially supported by the National Research Foundation of Korea (NRF) under Grant No. RS-2023-00222663 (Center for Optimizing Hyperscale AI Models and Platforms), and by the Institute for Information and Communications Technology Promotion (IITP) under Grant No. 2018-0-00581 (CUDA Programming Environment for FPGA Clusters) and No. RS-2025-02304554 (Efficient and Scalable Framework for AI Heterogeneous Cluster Systems), all funded by the Ministry of Science and ICT (MSIT) of Korea. Additional support was provided by the BK21 Plus Program for Innovative Data Science Talent Education (Department of Data Science, SNU, No. 5199990914569) and the BK21 FOUR Program for Intelligent Computing (Department of Computer Science and Engineering, SNU, No. 4199990214639), both funded by the Ministry of Education (MOE) of Korea. This work was also partially supported by the Artificial Intelligence Industrial Convergence Cluster Development Project, funded by the MSIT and Gwangju Metropolitan City. Research facilities were provided by ICT at Seoul National University.

%% file: sections/Appendix-Hyperparameters.tex
\input{sections/tables/table-hyperparameters}
\section{Hyperparameters and Details of Training \label{sec:hyperparameters}}

We use a total of 48 NVIDIA H100 GPUs for training models. For the detailed statistics and hyperparameters of the training, please refer to Table~\ref{tab:hyperparameters}.

%% file: sections/tables/table-hyperparameters.tex
\begin{table*}[htbp]
\centering

\resizebox{\textwidth}{!}{%
\begin{tabular}{@{}c|c|c|c|c|c|c|c|c@{}}
\toprule
\textbf{}                      & \textbf{Optimizer}                                                                                        & \textbf{Epochs} & \textbf{Tokens/epoch} & \textbf{Batch size} & \textbf{\begin{tabular}[c]{@{}c@{}}Sequence\\ Length\end{tabular}} & \textbf{\begin{tabular}[c]{@{}c@{}}Learning Rate\\ Scheduler\end{tabular}}                                 & \textbf{\begin{tabular}[c]{@{}c@{}}Peak\\ Learning rate\end{tabular}} & \textbf{\begin{tabular}[c]{@{}c@{}}Total\\ H100 hours\end{tabular}} \\ \midrule
Continual Pretraining          & \multirow{3}{*}{\begin{tabular}[c]{@{}c@{}}AdamW\\ $\beta=[0.9, 0.95]$\\ $\epsilon=10^{-5}$\end{tabular}} & 1               & 102B                  & 1104                & \multirow{3}{*}{8192}                                              & \multirow{3}{*}{\begin{tabular}[c]{@{}c@{}}Cosine Decay\\ Warmup steps=1\%\\ Decay ratio=0.1\end{tabular}} & 1.2e-4                                                                & 3,150                                                               \\
Supervised Fine-Tuning         &                                                                                                           & 3               & 99.4M                 & 32                  &                                                                    &                                                                                                            & 5.0e-6                                                                & 54.6                                                                \\
Direct Preference Optimization &                                                                                                           & 1               & 48.5M                 & 32                  &                                                                    &                                                                                                            & 5.0e-6                                                                & 6.9                                                                \\ \bottomrule
\end{tabular}%
}
\caption{Hyperparameters and details at each stage of training. For Direct Preference Optimization, we use $\gamma=0.5$ and $\beta=0.1$. \label{tab:hyperparameters}}
\end{table*}

%% file: sections/Appendix-Crawling.tex
\section{Crawling Rules}
\label{sec:crawling-rules}

We collect data from six major web services (Naver Blog~\footnote{\url{https://blog.naver.com/}}, Tistory Blog~\footnote{\url{https://tistory.com/}}, Naver Cafe~\footnote{\url{https://cafe.naver.com/}}, Daum Cafe~\footnote{\url{https://cafe.daum.net/}}, Naver News~\footnote{\url{https://news.naver.com/}}, Daum News~\footnote{\url{https://blog.daum.net/}}).  
All crawling rules are designed to meet the following goals:
\begin{itemize}
    \item{We do not collect articles with restricted visibility settings (e.g., posts available only to certain members) to ensure privacy and to comply with Korean laws.}
    \item{We do not send excessive load to the servers to ensure that our data collection does not interfere with the normal operation of the target websites.}
    \item{If a section of a website does not appear to contain meaningful text or appears to contain repetitive text, we skip collecting that section to efficiently gather useful Korean text.}
\end{itemize}

We implement measures to limit the load imposed on the web services by adjusting the interval between two consecutive requests sent by a single worker (i.e., a public IP) to a web service.  
The rules are as follows:
\begin{itemize}
    \item{We utilize a maximum of 20 public IPs to crawl a single web service}
    \item{The default interval between requests is set to 1 second}
    \item{If any request returns an HTTP response code 429 (Too Many Requests) or returns no response we abort the crawling process and manually check whether the target web service is operating normally.}
    \item{If a request returns an HTTP error code 403 or 404, we proceed to the next URL.}
    \item{If a request returns any other HTTP error code (e.g., 501), we increase the interval to 15 seconds.}
    \item{If consecutive requests to the same URL return HTTP error codes, we double the interval each time. For example, if three consecutive requests fail, we wait 60 seconds before the fourth attempt.}
    \item{If five consecutive requests fail, we proceed to the next URL.}
    \item{If a request returns a successful response (or a redirect), we reset the interval to the default.}
\end{itemize}

\subsection{Blog}

We collect only publicly accessible data from blog posts. We implement a filtering process to avoid collecting blogs primarily containing multimedia only. For each blog, we sample up to 500 recent posts. If more than 80\% of these sampled posts contain no text, we cease further collection from that blog.

\subsection{News}

We collect all the news articles that are open to the general public at the time of data collection.

\subsection{Café}

A Café (online community) is composed of multiple boards, each containing articles.
We do not access articles that are not allowed to disclosed to public.
To improve data collection efficiency, we avoid accessing boards that does not seem to contain text, seem to contain repetitive text, and mostly contains inaccessible articles.
\begin{itemize}
    \item {To improve the efficiency of data collection, we do not collect articles from special-purpose boards. Special-purpose boards serve functions unrelated to the delivery of Korean textual content. For example, Photo Galleries mainly contain multimedia resources such as images, while Memo Boards allow members to leave short messages to one another. Other special-purpose boards provide external links, visual separators between sections, or serve other non-textual functions. Since these boards do not typically contain high-quality Korean text, we exclude them from our crawling targets.}
    
    \item {Boards that are not accessible to the general public are excluded from data collection.}
    
    \item {We further exclude boards whose names contain certain keywords, in order to avoid collecting repetitive or low-information content. 
    For instance, boards with names including "가입인사" or "가입 인사" (First Greeting) are typically filled with introductory posts from new members, often consisting of repetitive greetings. 
    Boards named with variations of "출석"(Attendance), such as "출석 체크", "출첵", or "출석체크" (Attendance), usually consist of brief daily messages such as "좋은 아침입니다 (Good morning)" or "출석합니다 (I’m here)." Boards with keywords "등업" (Promotion) or "신청" (Application) often contain boilerplate requests specific to the operational rules of a particular Cafe.}
    
    \item {If a board is publicly accessible but most of its articles are not, we stop crawling the board. This decision is based on a threshold using an exponential moving average (EMA) of article accessibility. Specifically, we compute EMA as follows:
    \[
    x_t = 
    \begin{cases}
    1 & \text{if the } t\text{-th article is accessible} \\
    0 & \text{otherwise}
    \end{cases}
    \]
    \[
    \text{EMA}_0 = 1, \quad
    r = \frac{2}{n+1}, \quad n = 10
    \]
    \[
    \text{EMA}_{t+1} = (1 - r)\,\text{EMA}_t + r\,x_t
    \]
    For each board, we access its articles in reverse chronological order (most recent first): we stop crawling the board if its EMA drops below 0.15, under the assumption that the majority of the articles are restricted.}
\end{itemize}

%% file: sections/Appendix-Post-Training.tex
\section{Details on Post Training}
\label{sec:post-training-details}

\input{sections/tables/table-post-training-sources}


We provide a detailed explanation of the process used to construct the dataset for post-training. This includes the steps taken to collect, filter, and format the data to align with the specific requirements of SFT and DPO, ensuring the resulting dataset is suitable for enhancing downstream task performance.

\subsection{Question-and-answer formatting}
Here, we describe the question-and-answer format into which public datasets were converted for use in SFT and DPO. Below are descriptions of the converted results for each source dataset. For SFT, we used prompt–chosen response pairs as question-answer pairs. For DPO, we used prompt–chosen response–rejected response triplets if the rejected response is available.

\paragraph{Ending-completion type datasets.}
HellaSwag~\cite{zellers2019hellaswag}, KoBEST-HellaSwag~\cite{jang-etal-2022-kobest}, and OpenbookQA~\cite{mihaylov-etal-2018-suit} fall into this category. These datasets require selecting the most appropriate ending to complete a given context presented as an incomplete sentence. We structured the context as the prompt, the correct choice as the chosen response, and the incorrect choices as rejected responses. Table~\ref{tab:QAformatting-hellaswag} shows an example of the formatted result for HellaSwag. The others follow a similar format.

\paragraph{Fill-in-the-blank type dataset.}
WinoGrande~\cite{sakaguchi2021winogrande} falls into this category. These datasets present a sentence with a blank in the middle and require selecting the correct word to fill in the blank from two choices. Considering the autoregressive nature of LLMs, we constructed the prompt using the context up to the word immediately before the blank, and the response using the remaining part of the sentence including the blank. The chosen response is the phrase with the blank filled in using the correct choice, while the rejected response is the phrase with the blank filled in using the incorrect choice. Table~\ref{tab:QAformatting-winogrande} shows an example of the formatted result for WinoGrande.

\paragraph{MMLU-style datasets.}
ARC-E/C~\cite{clark2018think}, MMLU~\cite{hendrycks2021mmlu}, and KMMLU~\cite{son2024kmmlu} fall into this category. In these datasets, a question is presented along with four answer choices, and the task is to select the most appropriate one. We constructed the prompt by concatenating the question with each choice prefixed by its corresponding label (A, B, C, or D). The correct choice was used as the chosen response, while the incorrect choices were used as rejected responses. Table~\ref{tab:QAformatting-arc} shows an example of the formatted result for ARC-E. The others follow a similar format.

\paragraph{Math-problem-solving datasets.}
GSM8K~\cite{cobbe2021gsm8k} and OrcaMath~\cite{mitra2024orca} fall into this category. These are datasets where a math question is given, and the task is to generate a step-by-step solution along with the final answer. We used the question as the prompt and the full reasoning and answer as the response. In the GSM8K training set, equations are tagged with << >>, but since these tags are not essential to the model’s reasoning process, we removed them. Table~\ref{tab:QAformatting-gsm8k} shows an example of the formatted result for GSM8K. The other follow a similar format.

\paragraph{Coding-problem-solving datasets.}
MBPP~\cite{austin2021program} fall into this category. This dataset provide a question along with a corresponding code-based answer. Instead of using the entire training set, we selected only those examples that do not require class definitions and contain a single top-level function. The question was inserted as the docstring of the unique function. The prompt consists of the function definition including the docstring, while the response consists of the function body. Table~\ref{tab:QAformatting-mbpp} shows an example of the formatted result for MBPP.

\subsection{Construction of synthetic dataset}
As mentioned in \ref{para:synthetic-datasets}, we generated responses to high-quality prompts—collected online—for tasks such as math problem solving, instruction following, and coding, using open LLMs. These generated responses were used for fine-tuning. In this section, we describe the detailed process.

\paragraph{Collecting prompts.}
For each task, the authors manually selected open datasets that has high-quality questions. The specific datasets used for each task are as follows:
\begin{itemize}
    \item{For math problem solving, the English questions were taken from GSM8K and OrcaMath. The Korean questions were sourced from MWP\_KR\_DATA~\cite{MWP_KR_DATA} dataset and a subset of the publicly available HuggingFace dataset ChuGyouk/AI-MO-NuminaMath-CoT-Ko~\cite{numina_math_ko}, which provides Korean translations of various math benchmarks compiled in AI-MO/NuminaMath-CoT~\cite{numina_math_datasets}. Specifically, we selected instances whose source field corresponds to GSM8K or OrcaMath. These English and Korean questions were used directly without processing.}
    \item{For instruction following, the English questions were taken from SlimOrca~\cite{SlimOrca,mukherjee2023orca,longpre2023flan}, and the Korean ones from the KoAlpaca~\cite{alpaca,koalpaca} dataset. SlimOrca is a curated subset of OpenOrca~\cite{OpenOrca,mukherjee2023orca,longpre2023flan}, consisting of system prompts, user queries, and LLM responses. KoAlpaca is a dataset of user's questions and answers collected from Naver’s inter-user Q\&A platform, Naver Knowledge-iN. In both cases, user questions were extracted, and random instructions were appended to them to construct the prompts.}
    \item{For code problem solving, we used the questions of MBPP and LeetCodeDataset~\cite{xia2025leetcodedataset} in the formatted prompts as described in the table~\ref{tab:QAformatting-mbpp}.}
\end{itemize}

\paragraph{Generating responses.}
We generated synthetic answers to the collected prompts using large-scale language models (LLMs). Through multiple rounds of trial and error, we observed that the quality of LLM responses can vary significantly depending on the system prompt and whether few-shot examples are used. If few-shot examples are used, they are randomly sampled from the training set. The system prompts and the number of few-shot examples are described in the table~\ref{tab:System-prompt-etc}.
We used vLLM(v0.6.6)~\cite{kwon2023efficient} to efficiently generate a large number of responses. vLLM is a library optimized for fast and memory-efficient inference with large language models, making it well suited for large-scale generation tasks.
To ensure diversity in the generated responses, we employed multiple models. Specifically, Qwen2.5-Math-72B-Instruct was used for generating responses to English math questions, Qwen2.5-Coder-32B-Instruct for English coding questions, and Qwen2.5-32B-Instruct for English instruction-following questions. EXAONE-3.5-32B-Instruct was used to generate responses for all types of Korean questions, while Llama-3.3-70B-Instruct was used for both Korean and English questions across all task types. Each model was selected based on publicly available benchmark results and our practical experience with their response quality.
\paragraph{QA formatting.}
To use the generated responses for SFT and DPO datasets, they are classified into chosen and rejected responses, and then converted into the question-and-answer format as described above.


%% file: sections/tables/table-post-training-sources.tex
\begin{table*}[t]
\centering
\resizebox{\textwidth}{!}{%
\begin{tabular}{@{}cccccc@{}}
\toprule
\textbf{Language}        & \textbf{Domain}                        & \textbf{Type}                                                                                                             & \textbf{Source}                                                                                                         & \textbf{\# Data for SFT}                                                             & \textbf{\# Data for DPO}                                               \\ \midrule
\multirow{6}{*}{English} & Commonsense Reasoning                  & \begin{tabular}[c]{@{}c@{}}Training set\\ Training set\end{tabular}                                                       & \begin{tabular}[c]{@{}c@{}}HellaSwag\\ WinoGrande\end{tabular}                                                          & \begin{tabular}[c]{@{}c@{}}39905\\ 40398\end{tabular}                                & \begin{tabular}[c]{@{}c@{}}119715\\ 40398\end{tabular}                 \\ \cmidrule(l){2-6} 
                         & Reading Comprehension                  & Training set                                                                                                              & OBQA                                                                                                                    & 4957                                                                                 & 14871                                                                  \\ \cmidrule(l){2-6} 
                         & Knowledge                              & Training set                                                                                                              & MMLU                                                                                                                    & 99842                                                                                & 299526                                                                 \\ \cmidrule(l){2-6} 
                         & Math \& Science                        & \begin{tabular}[c]{@{}c@{}}Training set\\ Synthetic\\ Training set\\ Synthetic\\ Training set\\ Training set\end{tabular} & \begin{tabular}[c]{@{}c@{}}GSM8K\\ GSM8K\\ OrcaMath\\ OrcaMath \\ ARC-Easy\\ ARC-Challenge\end{tabular}                 & \begin{tabular}[c]{@{}c@{}}7473\\ 12363\\ 112743\\ 145179\\ 2251\\ 1119\end{tabular} & \begin{tabular}[c]{@{}c@{}}-\\ 2335\\ -\\ -\\ 6751\\ 3357\end{tabular} \\ \cmidrule(l){2-6} 
                         & Coding                                 & \begin{tabular}[c]{@{}c@{}}Synthetic\\ Synthetic\end{tabular}                                                             & \begin{tabular}[c]{@{}c@{}}MBPP dataset\\ LeetCode dataset\end{tabular}                                                 & \begin{tabular}[c]{@{}c@{}}1475\\ 2698\end{tabular}                                  & \begin{tabular}[c]{@{}c@{}}-\\ -\end{tabular}                          \\ \cmidrule(l){2-6} 
                         & Instruction Following                  & Synthetic                                                                                                                 & SlimOrca                                                                                                                & 319526                                                                               & 18889                                                                  \\ \midrule
\multirow{5}{*}{Korean}  & Commonsense Reasoning                  & Training set                                                                                                              & KoBEST-HellaSwag                                                                                                        & 2029                                                                                 & 6087                                                                   \\ \cmidrule(l){2-6} 
                         & Knowledge                              & Training set                                                                                                              & KMMLU                                                                                                                   & 208522                                                                               & 625566                                                                 \\ \cmidrule(l){2-6} 
                         & Math \& Science                        & \begin{tabular}[c]{@{}c@{}}Synthetic\\ Training set\\ Synthetic\\ Synthetic\end{tabular}                                  & \begin{tabular}[c]{@{}c@{}}Translated GSM8K\\ Translated OrcaMath\\ Translated OrcaMath\\  mwp-korean-2021\end{tabular} & \begin{tabular}[c]{@{}c@{}}11238\\ 112743\\ 133380\\ 3126\end{tabular}               & \begin{tabular}[c]{@{}c@{}}1804\\ -\\ -\\ 660\end{tabular}             \\ \cmidrule(l){2-6} 
                         & \multirow{1}{*}{Instruction Following} & Synthetic                                                                                                                 & KoAlpaca                                                                                                                & 49832                                                                                & 3825                                                                                                                                      \\ \bottomrule
\end{tabular}%
}
\caption{Data sources and distribution for post-training dataset. For explanation of the type of data sources, please refer to Section~\ref{sec:dataset-post-training}}
\label{tab:post-training-sources}
\end{table*}

%% file: sections/Appendix-Benchmark-Eval-Method.tex
\section{Evaluation Method of Downstream Benchmarks}
\label{sec:benchmark-eval-method}

To ensure all evaluation is performed with the same settings, we recompute numbers with our own evaluation pipeline, most of which originated from lm-evaluation-harness\cite{eval-harness}. We add our model implementation to its supported models. Table ~\ref{tab:benchmark-statistics} summarizes the benchmarks we used and the number of examples in benchmarks.

\input{sections/tables/table-benchmark-statistics}

\newcommand{\testsplit}[1]{We use the #1 split for evaluation.}

\newcommand{\mmlustyle}{We format questions and answers in MMLU-Style\cite{hendrycks2021mmlu}.}

\newcommand{\accnorm}{For each choice, we compute the negative log-likelihood of the ending tokens, normalized by the length of the ending. The choice with the highest normalized score is selected.}

\newcommand{\acc}{For each choice, we compute the negative log-likelihood of the ending tokens. The choice with the highest score is selected.}

\newcommand{\fone}{We compute the F1 score.}

\newcommand{\exactmatch}{We compute the exact match score.}

\newcommand{\instloose}{We compute instruction level loose accuracy.}

\newcommand{\passatone}{We report pass@1.}

\newcommand{\zeroshot}{We report the zero-shot accuracy for the task.}
\newcommand{\noexample}{We do not provide examples in the prompt.}

\newcommand{\fewshot}[1]{We report the #1-shot accuracy for the task.}

\newcommand{\fewshotsplit}[2]{}

\newcommand{\firstn}[2]{We select the first #2 examples from the #1 split.}

\newcommand{\randn}[2]{We randomly sample #2 examples from the #1 split.}

\newcommand{\avoidself}{If the test instance appears among the few-shot examples, we replace it by sampling an additional example.}

\newcommand{\refertoprompt}[1]{Table~\ref{#1} describes the evaluation prompts.}


\subsection{English Downstream Tasks}

\subsubsection{HellaSwag~\cite{zellers2019hellaswag}}
\testsplit{validation}~\zeroshot ~\accnorm 
~\refertoprompt{tab:eval-prompts-hellaswag}


\subsubsection{WinoGrande~\cite{sakaguchi2021winogrande}}
\testsplit{validation}~\fewshot{5}~\randn{train}{5}
Following lm-evaluation-harness\cite{eval-harness}, we construct two context parts, each replacing '\_' with the choice. Then, for each context, we compute the negative log-likelihood of the ending part. We choose the context that yields the highest score in its ending part as a model response.
~\refertoprompt{tab:eval-prompts-winogrande}

\subsubsection{ARC-E/C~\cite{clark2018think}}
\testsplit{test}~\fewshot{25}~\randn{train}{25}~\mmlustyle ~\acc
~\refertoprompt{tab:eval-prompts-arc}

\subsubsection{MMLU~\cite{hendrycks2021mmlu}}
\testsplit{test}~\fewshot{5}~\firstn{dev}{5}~\mmlustyle ~\acc
~\refertoprompt{tab:eval-prompts-mmlu}

\subsubsection{OpenbookQA~\cite{mihaylov-etal-2018-suit}}
\testsplit{test}~\zeroshot ~\acc
~\refertoprompt{tab:eval-prompts-openbookqa}

\subsubsection{GSM8K~\cite{cobbe2021gsm8k}}
\testsplit{test}~\fewshot{8} Following Llama 3\cite{grattafiori2024llama}, We use 8-shot examples described in ~\cite{wei2022chain}. ~\exactmatch
~\refertoprompt{tab:eval-prompts-gsm8k}

\subsubsection{IFEval~\cite{zhou2023instruction}}
\testsplit{test}~\zeroshot ~\instloose
~\refertoprompt{tab:eval-prompts-ifeval}

\subsubsection{HumanEval~\cite{chen2021humaneval}}
\testsplit{test}~\passatone ~\noexample
~\refertoprompt{tab:eval-prompts-humaneval}

\subsection{Korean Downstream Tasks}

\subsubsection{KoBEST-HellaSwag~\cite{jang-etal-2022-kobest}}
\testsplit{test}~\zeroshot ~\accnorm ~\refertoprompt{tab:eval-prompts-kobest-hellaswag}


\subsubsection{Ko-WinoGrande}
\testsplit{test}~\zeroshot ~ We use the same method to build prompts and choose answer choice as the case of WinoGrande. \refertoprompt{tab:eval-prompts-ko-winogrande}

\subsubsection{Ko-ARC-E/C}
\testsplit{test}~\fewshot{5}~\randn{test}{5}~\avoidself ~\accnorm ~\refertoprompt{tab:eval-prompts-ko-arc}

\subsubsection{KMMLU~\cite{son2024kmmlu}}
\testsplit{test}~\fewshot{5} We construct our context and endings in MMLU-Style except that we do not provide task description for each section. ~\randn{dev}{5}~\acc ~\refertoprompt{tab:eval-prompts-kmmlu}

\subsubsection{Ko-LAMBADA}
\testsplit{test}~\zeroshot ~\acc ~\refertoprompt{tab:eval-prompts-ko-lambada}

\subsubsection{Ko-GSM8K}
\testsplit{test}~\fewshot{5}~\randn{test}{5}\avoidself ~\exactmatch ~\refertoprompt{tab:eval-prompts-ko-gsm8k}

\subsubsection{Ko-IFEval}
\testsplit{test}~\zeroshot ~\instloose ~\refertoprompt{tab:eval-prompts-ko-ifeval}

\subsubsection{KR-HumanEval~\cite{kang2024krhumaneval}}
\testsplit{test}~\passatone ~\noexample ~\refertoprompt{tab:eval-prompts-kr-humaneval}



%% file: sections/tables/table-benchmark-statistics.tex
\begin{table*}[t]
\centering
\begin{tabular}{@{}c|c|c|c|c@{}}
\toprule
Language                    & Benchmark         & Train     & Validation    & Test  \\ \midrule
\multirow{10}{*}{English}   & HellaSwag         & 39,905    & 10,042        & 10,003\\
                            & WinoGrande        & 40,398    & 1,267         & 1,767 \\
                            & ARC Easy          & 2,251     & 570           & 2,376 \\
                            & ARC Challenge     & 1,119     & 299           & 1,172 \\
                            & MMLU              & 99,842    & 1,531         & 14,042\\
                            & OpenbookQA        & 4,957     & 500           & 500   \\
                            & GSM8K             & 7,473     & -         &  1,319     \\
                            & IFEval            & -         & -             & 541   \\
                            & HumanEval         & -         & -             & 164   \\ \midrule
\multirow{10}{*}{Korean}    & KoBEST-HellaSwag  & 2,029     & 500           & 500   \\
                            & Ko-WinoGrande     & -         & -             & 1,267 \\
                            & Ko-ARC Easy       & -         & -             & 2,376 \\
                            & Ko-ARC Challenge  & -         & -             & 1,167 \\
                            & KMMLU             & 208,522   & 225           & 35,030\\
                            & Ko-LAMBADA        & -         & -             & 2,255 \\
                            & Ko-GSM8K          & -         & -             & 1,319 \\
                            & Ko-IFEval         & -         & -             & 841   \\
                            & KR-HumanEval      & -         & -             & 164   \\ \bottomrule
\end{tabular}
\caption{Number of examples in benchmarks. - denotes 'Not Applicable(N/A)'}
\label{tab:benchmark-statistics}

\end{table*}

%% file: sections/Appendix-ARR-Checklist.tex
\section{Checklist for ARR Submission}

\label{sec:arr-checklist}

\paragraph{(A1) Limitations of the work}

Please refer to the Limitations section of the main text.

\paragraph{(A2) Potential risks of the work}
We crawl data from Naver, Daum, and Tistory, which already filters out the harmful contents, so additional process to remove offensive content is unnecessary.
We did not perform any anonymization or removal of personal information. We are currently conducting research on data anonymization.


\paragraph{(B1) Citations of the used artifacts}
We tried our best to cite all papers, code repositories, and resources we used in the main content of the paper.


\paragraph{(B2) License or terms of use of the artifacts} \mbox{}

\textbf{Crawled Data.}
Collecting a massive amount of webpages is a highly automated process, which makes it impractical to acquire explicit consents from every author of webpages. 
To ensure privacy and abide with relevant Korean Law, we do not collect webpages with restricted visibility settings at the time of accessing a webpage. 
We ensure our data collection do not disturb normal operation of the target website by adaptively changing the time interval between consecutive accesses to the same website, considering the error code that server returns.
We do not disclose collected web data to protect copyrights of authors.

\textbf{Data for Continual Pre-Training} For all data downloaded from AiHub, use of data for model training is explicitly permitted.
For data from KISTI, all data is explicitly licensed for non-commercial use. As an academic and non-profit institution, we utilized the data accordingly for non-commercial purposes.
RedPajama V2 and Dolma are licensed under Apache 2.0, and DCLM is licensed under the MIT license.

\textbf{Data for Post-Training} 
All datasets used for post-training data construction are appropriately licensed for research.
HellaSwag, MMLU, GSM8K, OrcaMath, and SlimOrca are licensed under MIT license.
ARC and KoBEST-HellaSwag are licensed under CC-BY-SA 4.0.
WinoGrande, OpenbookQA, AI-MO/NuminaMath-CoT, KoAlpaca, LeetCodeDataset, and MWP\_KR\_DATA are licensed under Apache 2.0. 
KMMLU is licensed under CC-BY-ND 4.0.
MBPP is licensed under CC-BY 4.0.
ChuGyouk/AI-MO-NuminaMath-CoT-Ko is licensed under CC-BY-NC 4.0.

\textbf{Benchmarks.}
We ensure all of benchmarks are properly licensed for public use and appropriate for evaluating language models. HellaSwag, MMLU, GSM8K, HumanEval, and KR-HumanEval are licensed under MIT license. 
ARC and KoBEST-HellaSwag are licensed under CC-BY-SA 4.0. 
WinoGrande, OpenbookQA, and IFEval are licensed under Apache 2.0. 
KMMLU is licensed under CC-BY-ND 4.0.

\textbf{Texts for constructing Korean benchmarks.}
We translate WinoGrande, ARC, GSM8K, IFEval to construct Ko-WinoGrande, Ko-ARC, Ko-GSM8K and Ko-IFEval, respectively.
Each benchmark's license explicitly permit to use, modify, and redistribute data under certain conditions.
See Benchmark paragraph for each benchmark's license.

\textbf{Texts for constructing Ko-LAMBADA benchmark.}
We construct the Ko-LAMBADA benchmark using only public domain texts, primarily classical literary works whose copyrights have expired. Therefore, there are no copyright issues. Since the content is fictional, the dataset does not contain personal information or real-world references that may raise ethical concerns.

\paragraph{(B3) Proper use of existing artifacts and Intended use of created artifacts} \mbox{}

\textbf{Benchmarks for evaluating this model.} 
For 9 benchmarks(HellaSwag, WinoGrande, ARC Easy/Challenge, MMLU, OpenbookQA, GSM8K, KoBEST-HellaSwag, and KMMLU), we utilize its train set as post-training data. 
For evaluating models, we utilize test set of each benchmark. If it is impossible to utilize test set for evaluation, we utilize validation set. For more detail, see Appendix \ref{sec:benchmark-eval-method}.

\textbf{Other academic benchmarks and open datasets.}
MBPP and NuminaMath-CoT are benchmarks for coding and math, respectively. We utilize only training set for construction of post-training data.
6 open datasets(KoAlpaca, LeetCodeDataset, OrcaMath, MWP\_KR\_DATA, ChuGyouk/AI-MO-NuminaMath-CoT-Ko, and SlimOrca) do not come with predefined train/validation splits, as they were originally released solely for training LM. In such cases, we used the available data exclusively for post-training purposes. 
No validation set or test set of any datasets was used during post-training.

\paragraph{(B4) Description of steps for removing personal identifiable information(PII) and offensive contents from data}

For offensive content, we rely on the fact that the collected web pages come from major web services that actively monitor and moderate user-generated content. Therefore, we assume that the collected data does not contain severely harmful material. In this work, we do not perform additional filtering of offensive content or removal of personally identifiable information (PII) during model training. However, we are actively working on data anonymization and related research to address these issues in future versions.

\paragraph{(B5) Documentation of the artifacts}
We will release the document along with the code after the review process is complete.

\paragraph{(B6) Statistics for the data} \mbox{}

\textbf{Continual Pre-Training} See Table \ref{tab:english-data-sources} and \ref{tab:korean-crawl-sources} for the dataset we used for continual pre-training. All of the data is used as the train split.

\textbf{Post-Training} See Table~\ref{tab:post-training-sources} for the dataset we used for post-training.

\textbf{Benchmark} For benchmarks, see Table~\ref{tab:benchmark-statistics}

\paragraph{(C1) Descriptions of the number of parameters in the model, the total computational budget, and computing infrastructure}

The number of parameters of the models is reported throughout the main text.
We use a total of 48 NVIDIA H100 GPUs. Please refer to Table~\ref{tab:hyperparameters} of Appendix~\ref{sec:hyperparameters} for the computational budget.

\paragraph{(C2) Details of experimental setup}
Experimental setups are explained throughout the main text.
Best-found hyperparameter values are described in Table~\ref{tab:hyperparameters} of Appendix~\ref{sec:hyperparameters}.

\paragraph{(C3) Descriptive statistics about results}
We report the single-run result as the model training requires a substantial amount of GPU hours in our computing budget. We could not run the same model training multiple times due to the limited resources.

\paragraph{(C4) Use of existing packages}\mbox{}

\textbf{Data Collection} We use beautifulsoup4 (v4.12.2)~\cite{richardson2007beautiful} to extract text from html. \\

\textbf{Model Training} We leverage PyTorch(v2.4.0)~\cite{paszke2019pytorch} as our primary deep-learning framework. We integrate DeepSpeed(v0.16.2)~\cite{rajbhandari2020zero} to optimize training. This includes using ZeRO stage 1 to enhance memory efficiency and enable distributed training. Additionally, we utilize TransformerEngine(v1.12)~\cite{nvidia_transformer_engine} for advanced optimizations specific to transformer architectures. This encompasses fused attention kernels and efficient FP8 matrix multiplication kernels. \\

\textbf{Evaluation}
We evaluate our models with our evaluation pipeline, most of which originated from lm-evaluation-harness~\cite{eval-harness}. We make minor modifications(e.g. modifying import paths) to execute evaluation codes within our codebase. We use vllm(v0.5.4)~\cite{kwon2023efficient} to parallelize and optimize model inference. Some downstream tasks require model outputs be evaluated with other packages except lm evaluaion harness. We use code evaluation functions of evaluate(v0.4.3)~\cite{von-werra-etal-2022-evaluate} for HumanEval/KR-HumanEval, language detection functions of langdetect(v1.0.9)~\cite{danilak2014langdetect} for IFEval/Ko-IFEval.


\paragraph{(D1) Full text of instructions or disclaimers of any risks}
See Table ~\ref{tab:instruction-translate-korean}, \ref{tab:instruction-translate-english} for instructions of translating English benchmarks. See Table ~\ref{tab:instruction-ko-lambada-korean}, ~\ref{tab:instruction-ko-lambada-english} for instructions of constructing Ko-LAMBADA.

\paragraph{(D2) Recruitment process and payment of paid participants}
The authors and the members of the institution involved in creating Korean benchmark.
We will acknowledge how they were supported by grants.
Their average monthly stipend ranges from 400 USD to 1,500 USD, and we allocated the workload accordingly to their stipend.
We acknowledged 10 USD worth of work per 1 hour's worth of work, which is considered reasonable considering the minimum wage in Korea.

\paragraph{(D3) Consent from the used/curated data}
The participants, including the authors, are the members of the research group. All participants were fully notified that their annotations will be used to construct a Korean Benchmark.

\paragraph{(D4) Review of data collection protocol by an ethics review board}
The dataset does not contain any content related to ethical issues.
The translated version is simply a Korean version of widely-used benchmarks for evaluating English language models.
Ko-LAMBADA is constructed with only public domain texts, primarily classical literary works whose copyrights have expired. 
Therefore, the dataset contains neither personal information nor real-world references that may raise ethical concerns, which makes the institutional reviewing process unnecessary.

\paragraph{(D5) Basic demographic and geographic characteristics of the annotator population}

All annotators are authors or the members of the research group. 15 Annotators are involved in creating the Korean benchmark. All annotators are Korean. All annotators are Asian, native to Korean, aged from 20 to 30 (adults).


\paragraph{(E1) Use of AI assistants}

We do not use AI assistants in our work.


%% file: sections/Appendix-Large-Tables.tex
\input{sections/tables/post_training/b1-01-hellaswag}
\input{sections/tables/post_training/b1-02-winogrande}
\input{sections/tables/post_training/b1-03-arc}
\input{sections/tables/post_training/b1-04-gsm8k}
\input{sections/tables/post_training/b1-05-mbpp}
\input{sections/tables/post_training/b2-01-system_prompt_few_shot_examples}

\input{sections/tables/eval-prompts/c1-01-hellaswag}
\input{sections/tables/eval-prompts/c1-03-winogrande}
\input{sections/tables/eval-prompts/c1-04-arc}
\input{sections/tables/eval-prompts/c1-05-mmlu}
\input{sections/tables/eval-prompts/c1-06-openbookqa}
\input{sections/tables/eval-prompts/c1-07-gsm8k}
\input{sections/tables/eval-prompts/c1-08-ifeval}
\input{sections/tables/eval-prompts/c1-09-humaneval}

\input{sections/tables/eval-prompts/c2-01-kobest-hellaswag}
\input{sections/tables/eval-prompts/c2-03-ko-winogrande}
\input{sections/tables/eval-prompts/c2-04-ko-arc}
\input{sections/tables/eval-prompts/c2-05-kmmlu}
\input{sections/tables/eval-prompts/c2-06-ko-lambada}
\input{sections/tables/eval-prompts/c2-07-ko-gsm8k}
\input{sections/tables/eval-prompts/c2-08-ko-ifeval}
\input{sections/tables/eval-prompts/c2-09-kr-humaneval}


\include{sections/tables/annotation-instruction/instruction-translation-korean}
\include{sections/tables/annotation-instruction/instruction-translation-english}
\include{sections/tables/annotation-instruction/instruction-lambaba-korean}
\include{sections/tables/annotation-instruction/instruction-lambada-english}

%% file: sections/tables/post_training/b1-01-hellaswag.tex

\begin{table*}[htbp]
\centering
\begin{tabular}{@{}p{0.10\textwidth}|p{0.90\textwidth}@{}}
\toprule

\textbf{Training set} & \tablecell{
\textbf{Context:} \\
Removing ice from car: Then, the man writes over the snow covering the window of a car, and a woman wearing winter clothes smiles. Then \\
\\
\textbf{Choices:} \\
- , the man adds wax to the windshield and cuts it.\\
- , a person board a ski lift, while two men supporting the head of the person wearing winter clothes snow as the we girls sled.\\
- , the man puts on a christmas coat, knitted with netting.\\
- , the man continues removing the snow on his car.\\
\\
\textbf{Answer:} \\
, the man continues removing the snow on his car.
}
\\ \midrule
 
\textbf{Prompt} & \tablecell{
Removing ice from car: Then, the man writes over the snow covering the window of a car, and a woman wearing winter clothes smiles. Then 
}    
\\ \midrule
 
\textbf{Chosen response} & \tablecell{
, the man continues removing the snow on his car.
}    
\\ \midrule

\textbf{Rejected responses} & \tablecell{
- , the man adds wax to the windshield and cuts it.\\
- , a person board a ski lift, while two men supporting the head of the person wearing winter clothes snow as the we girls sled.\\
- , the man puts on a christmas coat, knitted with netting.\\
}    

\\ \bottomrule
 
\end{tabular}%
\caption{Question-Answering formatting for HellaSwag}
\label{tab:QAformatting-hellaswag}
\end{table*}

%% file: sections/tables/post_training/b1-02-winogrande.tex
\begin{table*}[htbp]
\centering
\begin{tabular}{@{}p{0.10\textwidth}|p{0.90\textwidth}@{}}
\toprule

\textbf{Training set} & \tablecell{
\textbf{Sentence:} \\
Ian volunteered to eat Dennis's menudo after already having a bowl because \_ enjoyed eating intestine. \\
\\
\textbf{Choices:} \\
- Ian\\
- Dennis\\
\\
\textbf{Answer:} \\
Ian
}
\\ \midrule
 
\textbf{Prompt} & \tablecell{
Ian volunteered to eat Dennis's menudo after already having a bowl because 
}    
\\ \midrule

\textbf{Chosen response} & \tablecell{
Ian enjoyed eating intestine.
}    
\\ \midrule
 
\textbf{Rejected response} & \tablecell{
Dennis enjoyed eating intestine.
}
\\ \bottomrule
 
\end{tabular}%
\caption{Question-Answering formatting for WinoGrande}
\label{tab:QAformatting-winogrande}

\end{table*}

%% file: sections/tables/post_training/b1-03-arc.tex
\begin{table*}[htbp]
\centering
\begin{tabular}{@{}p{0.10\textwidth}|p{0.90\textwidth}@{}}
\toprule

\textbf{Training set} & \tablecell{
\textbf{Question:} \\
Which factor will most likely cause a person to develop a fever? \\
\\
\textbf{Choices:} \\
- a leg muscle relaxing after exercise\\
- a bacterial population in the bloodstream\\
- several viral particles on the skin\\
- carbohydrates being digested in the stomach \\
\\
\textbf{Answer:} \\
a bacterial population in the bloodstream
}
\\ \midrule
 
\textbf{Prompt} & \tablecell{
Question: Which factor will most likely cause a person to develop a fever? \\
A: a leg muscle relaxing after exercise\\
B: a bacterial population in the bloodstream\\
C: several viral particles on the skin\\
D: carbohydrates being digested in the stomach\\
Answer:
}    
\\ \midrule
 
\textbf{Chosen response} & \tablecell{
B: a bacterial population in the bloodstream
}    
\\ \midrule
 
\textbf{Rejected responses} & \tablecell{
- A: a leg muscle relaxing after exercise\\
- C: several viral particles on the skin\\
- D: carbohydrates being digested in the stomach\\
}
\\ \bottomrule
 
\end{tabular}%
\caption{Question-Answering formatting for ARC}
\label{tab:QAformatting-arc}

\end{table*}

%% file: sections/tables/post_training/b1-04-gsm8k.tex
\begin{table*}[htbp]
\centering
\begin{tabular}{@{}p{0.10\textwidth}|p{0.90\textwidth}@{}}
\toprule

\textbf{Training set} & \tablecell{
\textbf{Question:} \\
Natalia sold clips to 48 of her friends in April, and then she sold half as many clips in May. How many clips did Natalia sell altogether in April and May? \\
\\
\textbf{Answer:} \\
Natalia sold 48/2 = <<48/2=24>>24 clips in May.\\
Natalia sold 48+24 = <<48+24=72>>72 clips altogether in April and May.\\
\#\#\#\# 72
}
\\ \midrule
 
\textbf{Prompt} & \tablecell{
Q: Natalia sold clips to 48 of her friends in April, and then she sold half as many clips in May. How many clips did Natalia sell altogether in April and May?\\
A:
}    
\\ \midrule
 
\tablecell{\textbf{Chosen response}} & \tablecell{
Natalia sold 48/2 = 24 clips in May.\\
Natalia sold 48+24 = 72 clips altogether in April and May.\\
}
\\ \bottomrule
 
\end{tabular}%
\caption{Question-Answering formatting for GSM8K}
\label{tab:QAformatting-gsm8k}

\end{table*}

%% file: sections/tables/post_training/b1-05-mbpp.tex
\begin{table*}[htbp]
\centering
\begin{tabular}{@{}p{0.10\textwidth}|p{0.90\textwidth}@{}}
\toprule

\textbf{Training set} & \tablecell{
\textbf{Question:} \\
Write a python function to find the first repeated character in a given string. \\
\textbf{Code:} \\
def first\_repeated\_char(str1): \\
~~~~for index,c in enumerate(str1): \\
~~~~~~~~if str1[:index+1].count(c) > 1: \\
~~~~~~~~~~~~return c \\  
~~~~return "None"
}
\\ \midrule

\tablecell{\textbf{Prompt}} & \tablecell{
def first\_repeated\_char(str1): \\
~~~~""" \\
~~~~Write a python function to find the first repeated character in a given string. \\
~~~~"""
}
\\ \midrule

\tablecell{\textbf{Chosen response}} & \tablecell{
~~~~for index,c in enumerate(str1): \\
~~~~~~~~if str1[:index+1].count(c) > 1: \\
~~~~~~~~~~~~return c \\  
~~~~return "None" 
}
\\ \bottomrule
 
\end{tabular}%
\caption{Question-Answering formatting for MBPP}
\label{tab:QAformatting-mbpp}

\end{table*}

%% file: sections/tables/post_training/b2-01-system_prompt_few_shot_examples.tex
\begin{table*}[htbp]
\centering
\begin{tabular}{@{}p{0.10\textwidth}|p{0.90\textwidth}@{}}
\toprule

\textbf{Math problem solving (English)} & \tablecell{
\textbf{System prompt:} \\
Think carefully and reason through the following math problem to arrive at the answer. Write your answer as a number following 'The answer is'. Let's think step by step. \\
\\
\textbf{\# few-shots:} 5
}
\\ \midrule
 
\textbf{Math problem solving (Korean)} & \tablecell{
\textbf{System prompt:} \\
다음 수학 문제에 대해서, 충분히 생각하고 추론하여 답을 도출하세요. 답은 \#\#\#\# 뒤에 숫자로 쓰세요. \\
\\
\textbf{\# few-shots:} 3
}
\\ \midrule

\textbf{Code problem solving (English)} & \tablecell{
\textbf{System prompt:} \\
You are a helpful coding assistant. Your task is to complete Python function definitions that solve the given problem.\\
\\
You will be given:\\
- A Python function signature and docstring describing the expected behavior.\\
- You must implement the function body so that it matches the description.\\
- Do not write any test cases, print statements, or example usages. Only complete the function body.\\
- Return appropriate values as per the docstring.\\
\\
Be concise and correct. Assume all inputs are valid unless stated otherwise. \\
\\
\textbf{\# few-shots:} 0
}
\\ \midrule
 
\textbf{Instruction following (English)} & \tablecell{
\textbf{System prompt:} \\
Please answer the query strictly following the instruciton. \\
\\
\textbf{\# few-shots:} 0
}
\\ \midrule
 
\textbf{Instruction following (Korean)} & \tablecell{
\textbf{System prompt:} \\
지시사항에 충실히 따르면서 답변하세요. \\
\\
\textbf{\# few-shots:} 0
}
\\ \bottomrule
 
\end{tabular}%
\caption{System prompts and few-shot settings used for synthetic response generation}
\label{tab:System-prompt-etc}

\end{table*}

%% file: sections/tables/eval-prompts/c1-01-hellaswag.tex
\begin{table*}[htbp]
\centering
\begin{tabular}{@{}p{0.10\textwidth}|p{0.90\textwidth}@{}}
\toprule

\textbf{Problem} & \tablecell{
\textbf{Context:} \\
A man is being pulled on a water ski as he floats in the water casually. he \\
\\
\textbf{Choices:} \\
- mounts the water ski and tears through the water at fast speeds.\\
- goes over several speeds, trying to stay upright.\\
- struggles a little bit as he talks about it.\\
- is seated in a boat with three other people. \\
\\
\textbf{Answer:} \\
is seated in a boat with three other people.
}
\\ \midrule
 
\textbf{Context} & \tablecell{
A man is being pulled on a water ski as he floats in the water casually. he
}    
\\ \midrule
 
\textbf{Endings} & \tablecell{
mounts the water ski and tears through the water at fast speeds.
}
\\ \bottomrule
 
\end{tabular}%
\caption{Evaluation prompts for HellaSwag}
\label{tab:eval-prompts-hellaswag}

\end{table*}

%% file: sections/tables/eval-prompts/c1-03-winogrande.tex
\begin{table*}[htbp]
\centering
\begin{tabular}{@{}p{0.10\textwidth}|p{0.90\textwidth}@{}}
\toprule

\textbf{Problem} & \tablecell{
\textbf{Sentence:} \\
Sarah was a much better surgeon than Maria so \_ always got the easier cases. \\
\\
\textbf{Choices:} \\
- Sarah\\
- Maria\\
\\
\textbf{Answer:} \\
Maria
}
\\ \midrule
 
\textbf{Context1} & \tablecell{
The pants kept shrinking in the wash, but the socks weren't affected, because the pants were made of malleable material. \\\\
\textit{Few shot examples as above continue. Each example and the question instance is separated with two newline characters.}\\\\
Sarah was a much better surgeon than Maria so \textbf{Sarah}
}    
\\ \midrule

\textbf{Context2} & \tablecell{
The pants kept shrinking in the wash, but the socks weren't affected, because the pants were made of malleable material. \\\\
\textit{Few shot examples as above continue. Each example and the question instance is separated with two newline characters.}\\\\
Sarah was a much better surgeon than Maria so \textbf{Maria}
}    
\\ \midrule
 
\textbf{Ending} & \tablecell{
always got the easier cases.
}
\\ \bottomrule
 
\end{tabular}%
\caption{Evaluation prompts for WinoGrande}
\label{tab:eval-prompts-winogrande}

\end{table*}

%% file: sections/tables/eval-prompts/c1-04-arc.tex
\begin{table*}[htbp]
\centering
\begin{tabular}{@{}p{0.10\textwidth}|p{0.90\textwidth}@{}}
\toprule

\textbf{Problem} & \tablecell{
\textbf{Question:} \\
Which piece of safety equipment is used to keep mold spores from entering the respiratory system? \\
\\
\textbf{Choices:} \\
- safety goggles\\
- breathing mask\\
- rubber gloves\\
- lead apron \\
\\
\textbf{Answer:} \\
breathing mask
}
\\ \midrule
 
\textbf{Context} & \tablecell{
Question: Which of these resources will most likely be depleted first?\\
A: Wind\\
B: Solar energy\\
C: Fossil fuels\\
D: Water\\
Answer:  C\\\\
\textit{Few shot examples as above continue. Each example and the question instance is separated with two newline characters.} \\\\
Question: Which piece of safety equipment is used to keep mold spores from entering the respiratory system?\\
A: safety goggles\\
B: breathing mask\\
C: rubber gloves\\
D: lead apron\\
Answer:
}    
\\ \midrule
 
\textbf{Endings} & \tablecell{
A
}
\\ \bottomrule
 
\end{tabular}%
\caption{Evaluation prompts for ARC}
\label{tab:eval-prompts-arc}

\end{table*}

%% file: sections/tables/eval-prompts/c1-05-mmlu.tex
\begin{table*}[htbp]
\centering
\begin{tabular}{@{}p{0.10\textwidth}|p{0.90\textwidth}@{}}
\toprule

\textbf{Problem} & \tablecell{
\textbf{Question:} \\
Which of the following styles of fuzzer is more likely to explore paths covering every line of code in the following program? \\
\\
\textbf{Choices:} \\
- Generational\\
- Blackbox\\
- Whitebox\\
- Mutation-based\\
\\
\textbf{Answer:} \\
Whitebox
}
\\ \midrule
 
\textbf{Context} & \tablecell{
The following are multiple choice questions (with answers) about computer security.\\
SHA-1 has a message digest of\\
A. 160 bits\\
B. 512 bits\\
C. 628 bits\\
D. 820 bits\\
Answer: A\\\\
\textit{Few shot examples as above continue. Each example and the question instance is separated with two newline characters.} \\\\
Which of the following styles of fuzzer is more likely to explore paths covering every line of code in the following program?\\
A. Generational\\
B. Blackbox\\
C. Whitebox\\
D. Mutation-based\\
Answer: 
}    
\\ \midrule
 
\textbf{Endings} & \tablecell{A}
\\ \bottomrule
 
\end{tabular}%
\caption{Evaluation prompts for MMLU}
\label{tab:eval-prompts-mmlu}

\end{table*}

%% file: sections/tables/eval-prompts/c1-06-openbookqa.tex
\begin{table*}[htbp]
\centering
\begin{tabular}{@{}p{0.10\textwidth}|p{0.90\textwidth}@{}}
\toprule

\textbf{Problem} & \tablecell{
\textbf{Context:} \\
A person wants to start saving money so that they can afford a nice vacation at the end of the year. After looking over their budget and expenses, they decide the best way to save money is to \\
\\
\textbf{Choices:} \\
- make more phone calls\\
- quit eating lunch out\\
- buy less with monopoly money\\
- have lunch with friends\\
\\
\textbf{Answer:} \\
quit eating lunch out
}
\\ \midrule
 
\textbf{Context} & \tablecell{
A person wants to start saving money so that they can afford a nice vacation at the end of the year. After looking over their budget and expenses, they decide the best way to save money is to
}    
\\ \midrule
 
\textbf{Endings} & \tablecell{
make more phone calls
}
\\ \bottomrule
 
\end{tabular}%
\caption{Evaluation prompts for OpenbookQA}
\label{tab:eval-prompts-openbookqa}

\end{table*}

%% file: sections/tables/eval-prompts/c1-07-gsm8k.tex
\begin{table*}[htbp]
\centering
\begin{tabular}{@{}p{0.10\textwidth}|p{0.90\textwidth}@{}}
\toprule

\textbf{Problem} & \tablecell{
\textbf{Question:} \\
Janet’s ducks lay 16 eggs per day. She eats three for breakfast every morning and bakes muffins for her friends every day with four. She sells the remainder at the farmers' market daily for \$2 per fresh duck egg. How much in dollars does she make every day at the farmers' market? \\
\\
\textbf{Answer:} \\
Janet sells 16 - 3 - 4 = <{}<16-3-4=9>{}>9 duck eggs a day.\\
She makes 9 * 2 = \$<{}<9*2=18>{}>18 every day at the farmer’s market.\\
\textbf{\#\#\#\# 18}
}
\\ \midrule
 
\textbf{Prompt} & \tablecell{
Q: There are 15 trees in the grove. Grove workers will plant trees in the grove today. After they are done, there will be 21 trees. How many trees did the grove workers plant today?\\
A: There are 15 trees originally. Then there were 21 trees after some more were planted. So there must have been 21 - 15 = 6. The answer is 6.\\\\
\textit{Few shot examples as above continue. Each example and the question instance is separated with two newline characters.}\\\\
Q: Janet's ducks lay 16 eggs per day. She eats three for breakfast every morning and bakes muffins for her friends every day with four. She sells the remainder at the farmers' market daily for \$2 per fresh duck egg. How much in dollars does she make every day at the farmers' market?\\
A:
}    
\\ \midrule
 
\tablecell{\textbf{Expected} \\ \textbf{Output}} & \tablecell{
Janet starts with 16 eggs per day. She eats 3 eggs for breakfast, leaving her with 16 - 3 = 13 eggs for sale. She uses 4 eggs to bake muffins, leaving her with 13 - 4 = 9 eggs for sale at the market. At \$2 per egg, she makes 9 * \$2 = \$18 every day at the market. \textbf{The answer is 18}.
}
\\ \midrule
\tablecell{\textbf{Parsed} \\ \textbf{Answer}} & \tablecell{
18
}
\\ \bottomrule
 
\end{tabular}%
\caption{Evaluation prompts for GSM8K}
\label{tab:eval-prompts-gsm8k}

\end{table*}

%% file: sections/tables/eval-prompts/c1-08-ifeval.tex
\begin{table*}[htbp]
\centering
\begin{tabular}{@{}p{0.10\textwidth}|p{0.90\textwidth}@{}}
\toprule

\textbf{Problem} & \tablecell{
\textbf{Prompt:} \\
I am planning a trip to Japan, and I would like thee to write an itinerary for my journey in a Shakespearean style. You are not allowed to use any commas in your response. \\
\\
\textbf{Instruction ID List:} \\
punctuation\:no\_comma\\
}
\\ \midrule
 
\textbf{Prompt} & \tablecell{
I am planning a trip to Japan, and I would like thee to write an itinerary for my journey in a Shakespearean style. You are not allowed to use any commas in your response.
}    
\\ \midrule
 
\tablecell{\textbf{Expected} \\ \textbf{Output}} & \tablecell{
Act I Scene I\\\\
Thou art a traveler from afar seeking wonders untold\\
In land of rising sun where cherry blossoms gently unfold\\
Thou dost embark upon a quest both grand and bold\\
To witness temples ancient tales of old unfold(...\textit{abbreviated})
}
\\ \bottomrule
 
\end{tabular}%
\caption{Evaluation prompts for IFEval}
\label{tab:eval-prompts-ifeval}

\end{table*}

%% file: sections/tables/eval-prompts/c1-09-humaneval.tex
\begin{table*}[htbp]
\centering
\begin{tabular}{@{}p{0.10\textwidth}|p{0.90\textwidth}@{}}
\toprule

\textbf{Problem} & \tablecell{
\textbf{Prompt:} \\
def has\_close\_elements(numbers: List[float], threshold: float) -> bool: \\ 
~~~~""" Check if in given list of numbers, are any two numbers closer to each other than
given threshold. \\
~~~~>{}>{}> has\_close\_elements([1.0, 2.0, 3.0], 0.5) \\
~~~~False \\
~~~~>{}>{}> has\_close\_elements([1.0, 2.8, 3.0, 4.0, 5.0, 2.0], 0.3) \\
~~~~True \\
~~~~""" \\
\\
\textbf{Test Cases} \\
def check(candidate): \\
~~~~assert candidate([1.0, 2.0, 3.9, 4.0, 5.0, 2.2], 0.3) == True \\
~~~~assert candidate([1.0, 2.0, 3.9, 4.0, 5.0, 2.2], 0.05) == False \\
~~~~assert candidate([1.0, 2.0, 5.9, 4.0, 5.0], 0.95) == True \\
~~~~assert candidate([1.0, 2.0, 5.9, 4.0, 5.0], 0.8) == False \\
~~~~assert candidate([1.0, 2.0, 3.0, 4.0, 5.0, 2.0], 0.1) == True \\
~~~~assert candidate([1.1, 2.2, 3.1, 4.1, 5.1], 1.0) == True \\
~~~~assert candidate([1.1, 2.2, 3.1, 4.1, 5.1], 0.5) == False \\
}
\\ \midrule

\tablecell{\textbf{Model} \\ \textbf{Input}} & \tablecell{
def has\_close\_elements(numbers: List[float], threshold: float) -> bool: \\ 
~~~~""" Check if in given list of numbers, are any two numbers closer to each other than
given threshold. \\
~~~~>{}>{}> has\_close\_elements([1.0, 2.0, 3.0], 0.5) \\
~~~~False \\
~~~~>{}>{}> has\_close\_elements([1.0, 2.8, 3.0, 4.0, 5.0, 2.0], 0.3) \\
~~~~True \\
~~~~"""
}
\\ \midrule

\tablecell{\textbf{Sample} \\ \textbf{Output}} & \tablecell{
~~~~for idx, elem in enumerate(numbers): \\
~~~~~~~~for idx2, elem2 in enumerate(numbers): \\
~~~~~~~~~~~~if idx != idx2: \\
~~~~~~~~~~~~~~~~distance = abs(elem - elem2) \\
~~~~~~~~~~~~~~~~if distance < threshold: \\
~~~~~~~~~~~~~~~~~~~~return True \\
~~~~return False \\
}

\\ \bottomrule
 
\end{tabular}%
\caption{Evaluation prompts for HumanEval}
\label{tab:eval-prompts-humaneval}

\end{table*}

%% file: sections/tables/eval-prompts/c2-01-kobest-hellaswag.tex
\begin{table*}[htbp]
\centering
\begin{tabular}{@{}p{0.10\textwidth}|p{0.45\textwidth}|p{0.45\textwidth}@{}}
\toprule
 & Content in Korean & Content in English \\ \midrule
 
\textbf{Problem} & \tablecell{
\textbf{Context:} \\
엄마는 외출한 아들에게 메모지를 주며 장을 봐올 것을 부탁한다. 아들은 엄마의 심부름을 하기 위해 마트에 간다. 아들은 마트에서 카트를 챙기고 엄마가 준 메모지를 확인한다. \\
\\
\textbf{Choices:} \\
- 메모지를 확인하며 메모에 적힌 품목을 카트에 넣는다. \\
- 물건이 담긴 카트를 계산대에 가져간다. \\
- 아들이 계산이 완료된 물건을 장바구니에 담는다. \\
- 주인이 물건의 바코드를 찍고 카드를 받아 계산한다. \\
\\
\textbf{Answer:} \\
메모지를 확인하며 메모에 적힌 품목을 카트에 넣는다.
}
& \tablecell{
\textbf{Context:} \\
A mother gives her son a memo before going out and asks him to do the grocery shopping. The son goes to the mart to run the errand. At the mart, he gets a cart and checks the memo his mother gave him. \\
\\
\textbf{Choices:} \\
- He checks the memo and puts the listed items into the cart. \\
- He takes the cart with the items to the cashier. \\
- He puts the paid items into a shopping bag. \\
- The manager scans the items' barcodes and accepts the card for payment. \\
\\
\textbf{Answer:} \\
He checks the memo and puts the listed items into the cart.
}
\\ \midrule
 
\textbf{Context} & \tablecell{
엄마는 외출한 아들에게 메모지를 주며 장을 봐올 것을 부탁한다. 아들은 엄마의 심부름을 하기 위해 마트에 간다. 아들은 마트에서 카트를 챙기고 엄마가 준 메모지를 확인한다.
}
& \tablecell{
A mother gives her son a memo before going out and asks him to do the grocery shopping. The son goes to the mart to run the errand. At the mart, he gets a cart and checks the memo his mother gave him.
}

\\ \midrule
 
\textbf{Endings} & \tablecell{
메모지를 확인하며 메모에 적힌 품목을 카트에 넣는다.
}
& \tablecell{
He checks the memo and puts the listed items into the cart.
}
\\ \bottomrule
 
\end{tabular}%
\caption{Evaluation prompts for KoBest-Hellaswag}
\label{tab:eval-prompts-kobest-hellaswag}

\end{table*}

%% file: sections/tables/eval-prompts/c2-03-ko-winogrande.tex
\begin{table*}[htbp]
\centering
\begin{tabular}{@{}p{0.10\textwidth}|p{0.45\textwidth}|p{0.45\textwidth}@{}}
\toprule

 & Content in Korean & Content in English \\
\midrule

\textbf{Problem} & \tablecell{
\textbf{Sentence:} \\
지희는 채원이보다 훨씬 뛰어난 외과의사였기 때문에, \_는 항상 쉬운 케이스를 맡았습니다. \\
\\
\textbf{Choices:} \\
- 지희 \\
- 채원이 \\
\\
\textbf{Answer:} \\
채원이
} 
& \tablecell{
\textbf{Sentence:} \\
Jihee was a much better surgeon than Chaewon, so \_ always got the easier cases. \\
\\
\textbf{Choices:} \\
- Jihee \\
- Chaewon \\
\\
\textbf{Answer:} \\
Chaewon
}
\\ \midrule

\textbf{Context1} & \tablecell{
지희는 채원이보다 훨씬 뛰어난 외과의사였기 때문에, \textbf{지희}
}
& \tablecell{
Jihee was a much better surgeon than Chaewon, so \textbf{Jihee}
}
\\ \midrule

\textbf{Context2} & \tablecell{
지희는 채원이보다 훨씬 뛰어난 외과의사였기 때문에, \textbf{채원이}
}
& \tablecell{
Jihee was a much better surgeon than Chaewon, so \textbf{Chaewon}
}
\\ \midrule

\textbf{Ending} & \tablecell{
는 항상 쉬운 케이스를 맡았습니다.
}
& \tablecell{
always got the easier cases.
}
\\ \bottomrule

\end{tabular}%
\caption{Evaluation prompts for Ko-WinoGrande}
\label{tab:eval-prompts-ko-winogrande}
\end{table*}

%% file: sections/tables/eval-prompts/c2-04-ko-arc.tex
\begin{table*}[htbp]
\centering
\begin{tabular}{@{}p{0.10\textwidth}|p{0.45\textwidth}|p{0.45\textwidth}@{}}
\toprule
& Content in Korean & Content in English \\ \midrule
\textbf{Problem} 
& \tablecell{
\textbf{Question:} \\
곰팡이 포자가 호흡기로 들어가는 것을 막기 위해 사용되는 안전 장비에는 어떤 것이 있나요? \\
\\
\textbf{Choices:} \\
- 보안경\\
- 호흡 마스크\\
- 고무 장갑\\
- 납 앞치마\\
\\
\textbf{Answer:} \\
호흡 마스크
}
& \tablecell{
\textbf{Question:} \\
Which piece of safety equipment is used to keep mold spores from entering the respiratory system? \\
\\
\textbf{Choices:} \\
- safety goggles\\
- breathing mask\\
- rubber gloves\\
- lead apron \\
\\
\textbf{Answer:} \\
breathing mask
}
\\ \midrule
 
\textbf{Context} 
& \tablecell{
질문: 따오기는 한국의 습지에서 서식하던 조류로, 현재 멸종 위기에 처해 있습니다. 다음 중 따오기 멸종의 원인으로 가장 가능성이 높은 것은 무엇인가요?\\
답변: 과도한 사냥\\\\
\textit{Few shot examples as above continue. Each example and the question instance is separated with two newline characters.} \\\\
질문: 곰팡이 포자가 호흡기로 들어가는 것을 막기 위해 사용되는 안전 장비에는 어떤 것이 있나요?\\
답변:
}
& \tablecell{
Question: The crested ibis is a bird species that used to inhabit wetlands in Korea and is now facing extinction. Which of the following is the most likely cause of the crested ibis's extinction? \\
Answer: Overhunting\\\\
\textit{Few shot examples as above continue. Each example and the question instance is separated with two newline characters.} \\\\
Question: Which piece of safety equipment is used to keep mold spores from entering the respiratory system?\\
Answer:
}    
\\ \midrule
 
\textbf{Endings} 
& \tablecell{
호흡 마스크
}
& \tablecell{
breathing mask
}
\\ \bottomrule
 
\end{tabular}%
\caption{Evaluation prompts for Ko-ARC}
\label{tab:eval-prompts-ko-arc}

\end{table*}

%% file: sections/tables/eval-prompts/c2-05-kmmlu.tex
\begin{table*}[htbp]
\centering
\begin{tabular}{@{}p{0.10\textwidth}|p{0.45\textwidth}|p{0.45\textwidth}@{}}
\toprule
 & Content in Korean & Content in English \\ \midrule
 
\textbf{Problem} & \tablecell{
\textbf{Problem:} \\
TCP/IP 프로토콜 구조에 해당하지 않는 것은? \\
\textbf{Choices:} \\
- Network Access Layer \\
- Data Link Layer \\
- Physical Layer \\
- Transport Layer \\
\\
\textbf{Answer:} \\
Data Link Layer
}
& \tablecell{
\textbf{Problem:} \\
Which of the following does not belong to the TCP/IP protocol stack? \\
\textbf{Choices:} \\
- Network Access Layer \\
- Data Link Layer \\
- Physical Layer \\
- Transport Layer \\
\\
\textbf{Answer:} \\
Data Link Layer
}
\\ \midrule
 
\textbf{Context} & \tablecell{
DNS(Domain Name System) 서버 종류에 속하지 않는 것은?\\
A. Primary Server\\
B. Cache Server\\
C. Expert Server\\
D. Master Name Server\\
정답: C\\\\
\textit{Few shot examples as above continue. Each example and the question instance is separated with two newline characters.} \\\\
TCP/IP 프로토콜 구조에 해당하지 않는 것은? \\
A. Network Access Layer \\
B. Data Link Layer \\
C. Physical Layer \\
D. Transport Layer \\
정답: 
}
& \tablecell{
Which of the following is not a type of DNS (Domain Name System) server?\\
A. Primary Server\\
B. Cache Server\\
C. Expert Server\\
D. Master Name Server\\
Answer: C\\\\
\textit{Few shot examples as above continue. Each example and the question instance is separated with two newline characters.} \\\\
Which of the following does not belong to the TCP/IP protocol stack? \\
A. Network Access Layer \\
B. Data Link Layer \\
C. Physical Layer \\
D. Transport Layer \\
Answer:
}
\\ \midrule
 
\textbf{Endings} & \tablecell{
B
}
& \tablecell{
B
}
\\ \bottomrule
 
\end{tabular}%
\caption{Evaluation prompts for KMMLU}
\label{tab:eval-prompts-kmmlu}

\end{table*}

%% file: sections/tables/eval-prompts/c2-06-ko-lambada.tex
\begin{table*}[htbp]
\centering
\begin{tabular}{@{}p{0.10\textwidth}|p{0.45\textwidth}|p{0.45\textwidth}@{}}
\toprule
 & Content in Korean & Content in English \\ \midrule

\textbf{Problem} & \tablecell{
\textbf{Context:} \\
전차는 또 한 대 지나갔다. 승강대에 빈틈이 조금 있을 뿐, 미리 올라타서 가운데 숨어 있기에는 가장 적절한 전차였었다. 나는 혀를 한번 차고 담배를 꺼내어 붙여 물었다. 그 전차는 ○의 아내의 타는 정류장 앞에서 잠깐 멎었다가 다시 떠났다. 그러나 한 간을 나아가지 않아서 그 전차는 다시 멎었다. 나는 무심히 먹기 시작한 \_를 내어던지고 그편을 향하여 돌아섰다.\\
\\
\textbf{Choices:} \\
- 담배 \\
- 아내 \\
\\
\textbf{Answer:} \\
담배
} & \tablecell{
\textbf{Context:} \\
Another streetcar passed by. There was only a small space on the boarding platform, making it the most suitable car to board in advance and hide in the middle. I clicked my tongue and lit a cigarette. The streetcar briefly stopped in front of the station where ○'s wife usually boarded and then departed again. But it stopped again after moving just one section. I absentmindedly threw away the \_ I had started consuming and turned toward that side. \\
\\
\textbf{Choices:} \\
- cigarette \\
- wife \\
\\
\textbf{Answer:} \\
cigarette
}
\\ \midrule

\textbf{Context1} & \tablecell{
전차는 또 한 대 지나갔다. 승강대에 빈틈이 조금 있을 뿐, 미리 올라타서 가운데 숨어 있기에는 가장 적절한 전차였었다. 나는 혀를 한번 차고 담배를 꺼내어 붙여 물었다. 그 전차는 ○의 아내의 타는 정류장 앞에서 잠깐 멎었다가 다시 떠났다. 그러나 한 간을 나아가지 않아서 그 전차는 다시 멎었다. 나는 무심히 먹기 시작한 \textbf{담배}
} & \tablecell{
Another streetcar passed by. There was only a small space on the boarding platform, making it the most suitable car to board in advance and hide in the middle. I clicked my tongue and lit a cigarette. The streetcar briefly stopped in front of the station where ○'s wife usually boarded and then departed again. But it stopped again after moving just one section. I absentmindedly threw away the \textbf{cigarette}
}
\\ \midrule

\textbf{Context2} & \tablecell{
전차는 또 한 대 지나갔다. 승강대에 빈틈이 조금 있을 뿐, 미리 올라타서 가운데 숨어 있기에는 가장 적절한 전차였었다. 나는 혀를 한번 차고 담배를 꺼내어 붙여 물었다. 그 전차는 ○의 아내의 타는 정류장 앞에서 잠깐 멎었다가 다시 떠났다. 그러나 한 간을 나아가지 않아서 그 전차는 다시 멎었다. 나는 무심히 먹기 시작한 \textbf{아내}
} & \tablecell{
Another streetcar passed by. There was only a small space on the boarding platform, making it the most suitable car to board in advance and hide in the middle. I clicked my tongue and lit a cigarette. The streetcar briefly stopped in front of the station where ○'s wife usually boarded and then departed again. But it stopped again after moving just one section. I absentmindedly threw away the \textbf{wife}
}
\\ \midrule

\textbf{Endings} & \tablecell{
를 내어던지고 그편을 향하여 돌아섰다.
} & \tablecell{
I had started consuming and turned toward that side.
}
\\ \bottomrule

\end{tabular}%
\caption{Evaluation prompts for Ko-LAMBADA}
\label{tab:eval-prompts-ko-lambada}
\end{table*}

%% file: sections/tables/eval-prompts/c2-07-ko-gsm8k.tex
\begin{table*}[htbp]
\centering
\begin{tabular}{@{}p{0.10\textwidth}|p{0.45\textwidth}|p{0.45\textwidth}@{}}
\toprule

 & Content in Korean & Content in English \\
\midrule

\textbf{Problem} & \tablecell{
\textbf{Question:} \\
보리의 오리는 하루에 16개의 알을 낳습니다. 보리는 매일 아침 세 개를 아침 식사로 먹고, 네 개로 친구들을 위해 머핀을 만듭니다. 나머지는 매일 직거래 장터에서 신선한 오리알 한 개당 2000원에 판매합니다. 매일 직거래 장터에서 벌어들이는 돈은 몇 원일까요?\\
\\
\textbf{Answer:} \\
보리는 하루에 16 - 3 - 4 = <{}<16-3-4=9>{}>9개의 오리알을 판매합니다.\\
그녀는 매일 농산물 시장에서 9 * 2000 =<{}<9*2000=18000>{}>18000원을 벌고 있습니다.\\
\textbf{\#\#\#\# 18000}
} 
& \tablecell{
\textbf{Question:} \\
Bori’s ducks lay 16 eggs per day. She eats three for breakfast every morning and bakes muffins for her friends every day with four. She sells the remainder at the farmers' market daily for 2000 won per fresh duck egg. How much in won does she make every day at the farmers' market? \\
\\
\textbf{Answer:} \\
Bori sells 16 - 3 - 4 = <{}<16-3-4=9>{}>9 duck eggs a day.\\
She makes 9 * 2000 = <{}<9*2000=18000>{}>18000 won every day at the farmer’s market.\\
\textbf{\#\#\#\# 18000}
}
\\ \midrule

\textbf{Prompt} & \tablecell{
문제: 재하는 5일 동안 매일 자전거를 타고 출퇴근합니다. 그의 직장은 20km 떨어져 있습니다. 그는 또한 주말에 200km를 자전거로 타기도 합니다. 시속 25km로 자전거를 탈 수 있다면 일주일에 자전거를 타는 시간은 얼마나 될까요?\\
(\textit{Solution continues...})\\
따라서 그는 총 400/25=16시간을 탑니다.\\
\#\#\#\# 16\\\\
\textit{Few-shot examples continue above. Each example and the question are separated by two newline characters.}\\\\
문제: 보리의 오리는 하루에 16개의 알을 낳습니다. 보리는 매일 아침 세 개를 아침 식사로 먹고, 네 개로 친구들을 위해 머핀을 만듭니다. 나머지는 매일 직거래 장터에서 신선한 오리알 한 개당 2000원에 판매합니다. 매일 직거래 장터에서 벌어들이는 돈은 몇 원일까요?\\
답:
}
& \tablecell{
Problem: Jaeha commutes to work by bicycle for 5 days. His workplace is 20km away. He also rides 200km on weekends. If he rides at a speed of 25km/h, how many hours does he spend biking per week?\\
(\textit{Solution continues...})\\
So he spends a total of 400/25 = 16 hours.\\
\#\#\#\# 16\\\\
\textit{Few-shot examples continue above. Each example and the question are separated by two newline characters.}\\\\
Problem: Bori’s ducks lay 16 eggs per day. She eats three for breakfast every morning and bakes muffins for her friends every day with four. She sells the remainder at the farmers' market daily for 2000 won per fresh duck egg. How much in won does she make every day at the farmers' market?\\
Answer:
}
\\ \midrule

\tablecell{\textbf{Expected} \\ \textbf{Output}} & \tablecell{
보리의 오리는 하루에 16개의 알을 낳습니다.\\
보리는 매일 아침 세 개를 먹고, 네 개로 친구들을 위해 머핀을 만듭니다. 따라서 16 - 3 - 4 = 9개의 알이 남습니다.\\
매일 직거래 장터에서 신선한 오리알 한 개당 2000원에 판매하면, 9개의 알을 판매하면 9 * 2000 = 18000원을 벌 수 있습니다.\\
\textbf{\#\#\#\# 18000}
}
& \tablecell{
Bori’s ducks lay 16 eggs per day.\\
She eats 3 eggs for breakfast and uses 4 to make muffins for her friends. So, 16 - 3 - 4 = 9 eggs are left.\\
At the farmers' market, where each duck egg sells for 2000 won, she earns 9 * 2000 = 18000 won.\\
\textbf{\#\#\#\# 18000}
}
\\ \midrule

\tablecell{\textbf{Parsed} \\ \textbf{Answer}} & \tablecell{
18000
}
& \tablecell{
18000
}
\\ \bottomrule

\end{tabular}%
\caption{Evaluation prompts for Ko-GSM8K}
\label{tab:eval-prompts-ko-gsm8k}
\end{table*}

%% file: sections/tables/eval-prompts/c2-08-ko-ifeval.tex
\begin{table*}[htbp]
\centering
\begin{tabular}{@{}p{0.10\textwidth}|p{0.45\textwidth}|p{0.45\textwidth}@{}}
\toprule

 & Content in Korean & Content in English \\
\midrule

\textbf{Problem} & \tablecell{
\textbf{Prompt:} \\
인터넷의 역사와 인터넷이 우리 삶에 미친 영향에 대한 블로그 게시물을 작성하세요. 블로그 게시물은 청소년을 대상으로 작성하고 전체 답변을 큰따옴표로 감싸세요. \\
\\
\textbf{Instruction ID List:} \\
startend:quotation
} 
& \tablecell{
\textbf{Prompt:} \\
Write a blog post about the history of the internet and its impact on our lives. The blog post should be written for a teenage audience and the entire answer should be wrapped in double quotation marks. \\
\\
\textbf{Instruction ID List:} \\
startend:quotation
}
\\ \midrule

\textbf{Prompt} & \tablecell{
인터넷의 역사와 인터넷이 우리 삶에 미친 영향에 대한 블로그 게시물을 작성하세요. 블로그 게시물은 청소년을 대상으로 작성하고 전체 답변을 큰따옴표로 감싸세요.
}
& \tablecell{
Write a blog post about the history of the internet and its impact on our lives. The blog post should be written for a teenage audience and the entire answer should be wrapped in double quotation marks.
}
\\ \midrule

\tablecell{\textbf{Expected} \\ \textbf{Output}} & \tablecell{
"인터넷의 역사와 그 영향에 대해 알아보자! 인터넷은 우리가 오늘날 어떻게 소통하고 정보를 얻는지에 대한 근본적인 변화를 가져왔다. \\
(\textit{Content abbreviated...}) \\
인터넷은 단순히 기술적 발전을 넘어 우리 사회의 구조와 가치관까지 변화시킨 힘있는 힘이다. 앞으로도 계속 진화할 인터넷은 어떤 새로운 방식으로 우리 삶을 풍요롭게 만들어갈지 기대된다. 여러분의 생각은 어떤가? 어떻게 인터넷이 미래에 더 발전할 수 있을지 함께 고민해보자!"
}
& \tablecell{
"Let's learn about the history of the internet and its impact! The internet has fundamentally changed how we communicate and access information today. \\
(\textit{Content abbreviated...}) \\
The internet is not just a technological advancement, but a powerful force that has reshaped the structure and values of our society. As the internet continues to evolve, we can look forward to how it will enrich our lives in new ways. What do you think? How can the internet grow even more in the future? Let’s think about it together!"
}
\\ \bottomrule

\end{tabular}%
\caption{Evaluation prompts for Ko-IFEval}
\label{tab:eval-prompts-ko-ifeval}
\end{table*}

%% file: sections/tables/eval-prompts/c2-09-kr-humaneval.tex
\begin{table*}[htbp]
\centering
\begin{tabular}{@{}p{0.10\textwidth}|p{0.90\textwidth}@{}}
\toprule

\textbf{Problem} & \tablecell{
\textbf{Prompt:} \\
def has\_close\_elements(numbers: List[float], threshold: float) -> bool: \\ 
~~~~""" 주어진 숫자 배열에서 주어진 임계값보다 서로 가까운 두 숫자가 있는지 확인합니다. \\
~~~~>{}>{}> has\_close\_elements([1.0, 2.0, 3.0], 0.5) \\
~~~~False \\
~~~~>{}>{}> has\_close\_elements([1.0, 2.8, 3.0, 4.0, 5.0, 2.0], 0.3) \\
~~~~True \\
~~~~""" \\
\\
\textbf{Test Cases} \\
def check(candidate): \\
~~~~assert candidate([1.0, 2.0, 3.9, 4.0, 5.0, 2.2], 0.3) == True \\
~~~~assert candidate([1.0, 2.0, 3.9, 4.0, 5.0, 2.2], 0.05) == False \\
~~~~assert candidate([1.0, 2.0, 5.9, 4.0, 5.0], 0.95) == True \\
~~~~assert candidate([1.0, 2.0, 5.9, 4.0, 5.0], 0.8) == False \\
~~~~assert candidate([1.0, 2.0, 3.0, 4.0, 5.0, 2.0], 0.1) == True \\
~~~~assert candidate([1.1, 2.2, 3.1, 4.1, 5.1], 1.0) == True \\
~~~~assert candidate([1.1, 2.2, 3.1, 4.1, 5.1], 0.5) == False \\
}
\\ \midrule

\tablecell{\textbf{Docstring} \\(English)} & \tablecell{
Check if in given list of numbers, are any two numbers closer to each other than
given threshold.
}
\\ \midrule
 
\tablecell{\textbf{Model} \\ \textbf{Input}} & \tablecell{
def has\_close\_elements(numbers: List[float], threshold: float) -> bool: \\ 
~~~~""" 주어진 숫자 배열에서 주어진 임계값보다 서로 가까운 두 숫자가 있는지 확인합니다. \\
~~~~>{}>{}> has\_close\_elements([1.0, 2.0, 3.0], 0.5) \\
~~~~False \\
~~~~>{}>{}> has\_close\_elements([1.0, 2.8, 3.0, 4.0, 5.0, 2.0], 0.3) \\
~~~~True \\
~~~~"""
}
\\ \midrule

\tablecell{\textbf{Sample} \\ \textbf{Output}} & \tablecell{
~~~~for idx, elem in enumerate(numbers): \\
~~~~~~~~for idx2, elem2 in enumerate(numbers): \\
~~~~~~~~~~~~if idx != idx2: \\
~~~~~~~~~~~~~~~~distance = abs(elem - elem2) \\
~~~~~~~~~~~~~~~~if distance < threshold: \\
~~~~~~~~~~~~~~~~~~~~return True \\
~~~~return False \\
}

\\ \bottomrule
 
\end{tabular}%
\caption{Evaluation prompts for KR-HumanEval}
\label{tab:eval-prompts-kr-humaneval}

\end{table*}

%% file: sections/tables/annotation-instruction/instruction-translation-korean.tex
\begin{table*}[htbp]
\centering
\begin{tabular}{@{}p{0.12\textwidth}|p{0.88\textwidth}@{}}
\toprule

\textbf{작업 개요} & \tablecell{
초벌 번역: Deepl 번역 툴 사용하여 구축함 \\\\
작업 단계:
\begin{itemize}
    \item{1. 1차 번역 검수: 자동 번역 결과 확인 및 수정}
    \item{2. 1차 재검수: 본인이 검수한 내용 다시 확인}
    \item{3. 2차 재검수: 타인의 작업물 교차 검수}
\end{itemize}
추가 작업: 한글 기반 평가 코드 작성, 새로운 데이터 인스턴스 추가 등 (별도 담당자)
}

\\ \midrule
\tablecell{\textbf{번역 검수}\\\textbf{(1차 작업)}} & \tablecell{
목적: 자동 번역 결과에서 오류를 바로잡고 자연스러운 한국어 문장으로 다듬기 \\\\
작업 항목:
\begin{itemize}
    \item{영어 용어/고유명사를 한국식 명칭으로 수정}
    \item{번역이 비어 있거나 영어 원문이 그대로 남은 경우 직접 번역}
    \item{지나친 직역투/기계 번역 어투 수정}
    \item{문맥 흐름 확인 및 보완: 문장 연결이 어색하거나 번역 맥락이 어긋난 부분 수정}
\end{itemize}
}
\\ \midrule

\tablecell{\textbf{1차 재검수}\\\textbf{(자기 검수)}} & \tablecell{
목적: 자신이 작업한 번역 검수 내용을 다시 처음부터 확인하고 누락 및 오류 수정 \\\\
작업 항목:
\begin{itemize}
    \item{오탈자, 맞춤법, 띄어쓰기 오류 수정}
    \item{문화적 맥락 고려: 한국 문화와 맞지 않는 개념은 각색 또는 설명 보완}
    \item{어투 통일: 하나의 문단/예제 내에 어투 혼용 금지}
        \subitem{예: "했다", "했습니다", "했어요" -> "했다" 등으로 통일}
    \item{작업자 간 스타일 일관성 유지: 동일 표현은 전체 작업 내에서 동일하게 번역}
    \item{남아있는 번역투 재확인}
\end{itemize}
}
\\ \midrule

\tablecell{\textbf{2차 재검수}\\\textbf{(교차 검수)}} & \tablecell{
목적: 타인의 작업물을 검토하여 본인이 발견하지 못했던 오류를 보완 \\\\
작업 항목:
\begin{itemize}
    \item{전체내용 처음부터 읽으며 수정 필요 여부 판단}
    \item{표현 반복, 오탈자, 어색한 어순 등 수정}
    \item{표현 일관성 검토}
        \subitem{예: "했다", "했습니다", "했어요" -> "했다" 등으로 통일}
    \item{수정 시 간단한 주석 혹은 변경 이유 남기기: 공동 작업 시 이력 확인을 위한 메모 필요}
\end{itemize}
}
\\ \bottomrule
 
\end{tabular}%
\caption{Annotation Instructions for Translating Dataset}
\label{tab:instruction-translate-korean}
\end{table*}

%% file: sections/tables/annotation-instruction/instruction-translation-english.tex
\begin{table*}[htbp]
\centering
\begin{tabular}{@{}p{0.12\textwidth}|p{0.88\textwidth}@{}}
\toprule

\textbf{Task Overview} & \tablecell{
Initial translation: We used machine translation tools such as DeepL. \\
Task steps:
\begin{itemize}
    \item{1. First Review: Check and revise machine-translated results.}
    \item{2. Self-Review: Re-check your own revisions.}
    \item{3. Peer Review: Cross-check another person's work.}
\end{itemize}
Additional tasks: Writing evaluation code based on Korean, adding new data instances, etc. (assigned separately)
}

\\ \midrule
\tablecell{\textbf{First Review}\\\textbf{(Initial Task)}} & \tablecell{
Objective: Correct errors in machine-translated results and refine them into natural Korean sentences \\
Task items:
\begin{itemize}
    \item{Correct English terms/proper nouns into appropriate Korean equivalents.}
    \item{Manually translate any untranslated or English-only segments.}
    \item{Fix overly literal or unnatural machine-generated phrasing.}
    \item{Ensure logical flow: Fix awkward sentence transitions or context mismatches.}
\end{itemize}
}
\\ \midrule

\tablecell{\textbf{Self-Review}\\\textbf{(Secondary Check)}} & \tablecell{
Objective: Recheck your own revisions from the beginning to catch omissions and errors \\
Task items:
\begin{itemize}
    \item{Correct typos, spelling, and spacing errors.}
    \item{Consider cultural context: Adapt or clarify concepts unfamiliar in Korean culture.}
    \item{Unify tone: Avoid mixed tones within a paragraph/example.}
        \subitem{E.g., unify “did”, “has done”, “was doing” into one consistent form}
    \item{Maintain consistency across annotators: Use the same translation for the same expressions throughout the dataset.}
    \item{Double-check for remaining literal translations.}
\end{itemize}
}
\\ \midrule

\tablecell{\textbf{Peer Review}\\\textbf{(Cross Review)}} & \tablecell{
Objective: Review another person’s work to catch issues that may have been missed \\
Task items:
\begin{itemize}
    \item{Read through the content from the beginning and assess the need for revisions.}
    \item{Correct redundant expressions, typos, and awkward word orders.}
    \item{Use consistent expressions.}
        \subitem{E.g., unify "했다", "했습니다", "했어요" to "했다". }
        \subitem{(\textit{All of them are conjugated forms of 'did'})}
    \item{Leave brief comments or reasons for changes: helpful for tracking edits in collaborative work}
\end{itemize}
}
\\ \bottomrule

\end{tabular}%
\caption{Annotation Instructions for Translating Dataset(Translated to English)}
\label{tab:instruction-translate-english}
\end{table*}

%% file: sections/tables/annotation-instruction/instruction-lambaba-korean.tex
\begin{table*}[htbp]
\centering
\begin{tabular}{@{}p{0.20\textwidth}|p{0.80\textwidth}@{}}
\toprule

\textbf{데이터 교정} & \textbf{세부 내용} \\ \midrule

\textbf{언어적 자연스러움 검토} & \tablecell{
빈칸의 앞뒤 조사 및 문맥 흐름을 고려하여 자연스럽지 않은 \texttt{answer}나 \texttt{candidate}는 수정함. \\
다음과 같은 경우에는 수정 또는 제거:
\begin{itemize}
    \item \texttt{answer}와 \texttt{candidate}가 모두 정답으로 사용 가능해 혼란을 유발하는 경우
    \item \texttt{answer}를 앞 문맥에서 유추할 수 없는 경우
\end{itemize}
}
\\ \midrule

\textbf{통일성 확보} & \tablecell{
어투, 서술 방식, 표현 방식 등을 전체 데이터셋에 걸쳐 통일.\\
오탈자 및 논리적 오류도 함께 수정함.
}
\\ \midrule

\textbf{검토 및 품질 관리} & \textbf{세부 내용} \\ \midrule

\textbf{교차 검토} & \tablecell{
최소 1인 이상의 별도 검토자가 전체 데이터를 교차 검토함.\\
오류, 중복, 부자연스러운 문장 등을 최종적으로 점검하여 품질을 보장함.
}
\\ \midrule

\textbf{예시 구조} &  
\texttt{\{ "text": "조정은 늘 재정이 군색하였다... 조정의 \_은 더욱 군색하였다.", "answer": "재정", "candidate": "일문" \}} 
\\ \midrule

\textbf{유의 사항} & \tablecell{
\begin{itemize}
    \item 문맥 기반 추론이 가능한 문장을 선정할 것
    \item 일반 상식이나 배경지식이 아닌 텍스트 내 단서로부터 유추 가능해야 함
    \item 후보 문장이 너무 짧거나 단서가 부족한 문장은 제외
    \item 데이터셋 전체에 걸쳐 구성, 구조, 형식의 일관성 유지
\end{itemize}
}
\\ \bottomrule

\end{tabular}
\caption{Annotation Instructions for Ko-LAMBADA}
\label{tab:instruction-ko-lambada-korean}
\end{table*}

%% file: sections/tables/annotation-instruction/instruction-lambada-english.tex
\begin{table*}[htbp]
\centering
\begin{tabular}{@{}p{0.20\textwidth}|p{0.80\textwidth}@{}}
\toprule

\textbf{Data Correction} & \textbf{Details} \\ \midrule

\textbf{Fluency Check} & \tablecell{
Ensure linguistic naturalness. Modify any unnatural \texttt{answer} or \texttt{candidate} based on the context and surrounding particles. \\
Remove or revise if:
\begin{itemize}
    \item Both \texttt{answer} and \texttt{candidate} could plausibly be correct and cause confusion.
    \item The \texttt{answer} cannot be inferred from preceding context.
\end{itemize}
}
\\ \midrule

\textbf{Consistency} & \tablecell{
Unify tone, narrative style, and expression formats throughout the dataset. \\
Also, correct typographical and logical errors.
}
\\ \midrule

\textbf{Quality Assuarance} & \textbf{Details} \\ \midrule
\textbf{Cross-Verification} & \tablecell{
At least one additional reviewer must check the entire dataset. \\
Final checks should identify and fix errors, duplications, and unnatural sentences.
}
\\ \midrule

\textbf{Format Example} & 
\texttt{\{ "text": "조정은 늘 재정이 군색하였다... 조정의 \_은 더욱 군색하였다.", "answer": "재정", "candidate": "일문" \}} 
\\ \midrule

\textbf{Important Notes} & 
\begin{itemize}
    \item Choose sentences where the correct \texttt{answer} can be inferred from the given context.
    \item Avoid relying on general knowledge or background knowledge.
    \item Exclude overly short candidates or sentences lacking sufficient clues.
    \item Maintain consistency in structure, format, and composition across the entire dataset.
\end{itemize}

\\ \bottomrule

\end{tabular}
\caption{Annotation Instructions for Ko-LAMBADA(Tranlated to English)}
\label{tab:instruction-ko-lambada-english}
\end{table*}

%% file: main.bbl
\begin{thebibliography}{77}
\providecommand{\natexlab}[1]{#1}

\bibitem[{Austin et~al.(2021)Austin, Odena, Nye, Bosma, Michalewski, Dohan, Jiang, Cai, Terry, Le et~al.}]{austin2021program}
Jacob Austin, Augustus Odena, Maxwell Nye, Maarten Bosma, Henryk Michalewski, David Dohan, Ellen Jiang, Carrie Cai, Michael Terry, Quoc Le, and 1 others. 2021.
\newblock Program synthesis with large language models.
\newblock \emph{arXiv preprint arXiv:2108.07732}.

\bibitem[{Bak et~al.(2025)Bak, Lee, Ryu, Ham, Jung, Nam, Eo, Lee, Jung, Kim et~al.}]{bak2025kanana}
Yunju Bak, Hojin Lee, Minho Ryu, Jiyeon Ham, Seungjae Jung, Daniel~Wontae Nam, Taegyeong Eo, Donghun Lee, Doohae Jung, Boseop Kim, and 1 others. 2025.
\newblock Kanana: Compute-efficient bilingual language models.
\newblock \emph{arXiv preprint arXiv:2502.18934}.

\bibitem[{Beomi(2023)}]{koalpaca}
Beomi. 2023.
\newblock Koalpaca: Korean alpaca model based on stanford alpaca (feat. llama and polyglot-ko).
\newblock \url{https://github.com/Beomi/KoAlpaca}.

\bibitem[{Chen et~al.(2021)Chen, Tworek, Jun, Yuan, Pinto, Kaplan, Edwards, Burda, Joseph, Brockman et~al.}]{chen2021humaneval}
Mark Chen, Jerry Tworek, Heewoo Jun, Qiming Yuan, Henrique Ponde De~Oliveira Pinto, Jared Kaplan, Harri Edwards, Yuri Burda, Nicholas Joseph, Greg Brockman, and 1 others. 2021.
\newblock Evaluating large language models trained on code.
\newblock \emph{arXiv preprint arXiv:2107.03374}.

\bibitem[{Choi et~al.(2024{\natexlab{a}})Choi, Jeong, Park, Won, Lim, Kim, Kang, Yoon, Park, Lee, Lee, Hahm, Kim, and Lim}]{choi-etal-2024-optimizing}
ChangSu Choi, Yongbin Jeong, Seoyoon Park, Inho Won, HyeonSeok Lim, SangMin Kim, Yejee Kang, Chanhyuk Yoon, Jaewan Park, Yiseul Lee, HyeJin Lee, Younggyun Hahm, Hansaem Kim, and KyungTae Lim. 2024{\natexlab{a}}.
\newblock \href {https://aclanthology.org/2024.lrec-main.1095/} {Optimizing language augmentation for multilingual large language models: A case study on {K}orean}.
\newblock In \emph{Proceedings of the 2024 Joint International Conference on Computational Linguistics, Language Resources and Evaluation (LREC-COLING 2024)}, pages 12514--12526, Torino, Italia. ELRA and ICCL.

\bibitem[{Choi et~al.(2024{\natexlab{b}})Choi, Jeong, Park, Won, Lim, Kim, Kang, Yoon, Park, Lee et~al.}]{choi2024optimizing}
ChangSu Choi, Yongbin Jeong, Seoyoon Park, InHo Won, HyeonSeok Lim, SangMin Kim, Yejee Kang, Chanhyuk Yoon, Jaewan Park, Yiseul Lee, and 1 others. 2024{\natexlab{b}}.
\newblock Optimizing language augmentation for multilingual large language models: A case study on korean.
\newblock \emph{arXiv preprint arXiv:2403.10882}.

\bibitem[{ChuGyouk(2024)}]{numina_math_ko}
ChuGyouk. 2024.
\newblock Numinamath cot korean.
\newblock \url{https://huggingface.co/datasets/ChuGyouk/AI-MO-NuminaMath-CoT-Ko}.

\bibitem[{Clark et~al.(2018)Clark, Cowhey, Etzioni, Khot, Sabharwal, Schoenick, and Tafjord}]{clark2018think}
Peter Clark, Isaac Cowhey, Oren Etzioni, Tushar Khot, Ashish Sabharwal, Carissa Schoenick, and Oyvind Tafjord. 2018.
\newblock Think you have solved question answering? try arc, the ai2 reasoning challenge.
\newblock \emph{arXiv preprint arXiv:1803.05457}.

\bibitem[{Cobbe et~al.(2021{\natexlab{a}})Cobbe, Kosaraju, Bavarian, Chen, Jun, Kaiser, Plappert, Tworek, Hilton, Nakano, Hesse, and Schulman}]{cobbe2021gsm8k}
Karl Cobbe, Vineet Kosaraju, Mohammad Bavarian, Mark Chen, Heewoo Jun, Lukasz Kaiser, Matthias Plappert, Jerry Tworek, Jacob Hilton, Reiichiro Nakano, Christopher Hesse, and John Schulman. 2021{\natexlab{a}}.
\newblock Training verifiers to solve math word problems.
\newblock \emph{arXiv preprint arXiv:2110.14168}.

\bibitem[{Cobbe et~al.(2021{\natexlab{b}})Cobbe, Kosaraju, Bavarian, Chen, Jun, Kaiser, Plappert, Tworek, Hilton, Nakano et~al.}]{cobbe2021training}
Karl Cobbe, Vineet Kosaraju, Mohammad Bavarian, Mark Chen, Heewoo Jun, Lukasz Kaiser, Matthias Plappert, Jerry Tworek, Jacob Hilton, Reiichiro Nakano, and 1 others. 2021{\natexlab{b}}.
\newblock Training verifiers to solve math word problems.
\newblock \emph{arXiv preprint arXiv:2110.14168}.

\bibitem[{Cui et~al.(2023)Cui, Yang, and Yao}]{cui2023efficient}
Yiming Cui, Ziqing Yang, and Xin Yao. 2023.
\newblock Efficient and effective text encoding for chinese llama and alpaca.
\newblock \emph{arXiv preprint arXiv:2304.08177}.

\bibitem[{Danilák(2013)}]{danilak2014langdetect}
Michal Danilák. 2013.
\newblock \href {https://github.com/Mimino666/langdetect} {Langdetect}.
\newblock \emph{May}.

\bibitem[{Dou et~al.(2024{\natexlab{a}})Dou, Liu, Zeng, Guo, Zhou, Lu, and Lin}]{dou2024sailor}
Longxu Dou, Qian Liu, Guangtao Zeng, Jia Guo, Jiahui Zhou, Wei Lu, and Min Lin. 2024{\natexlab{a}}.
\newblock Sailor: Open language models for south-east asia.
\newblock \emph{arXiv preprint arXiv:2404.03608}.

\bibitem[{Dou et~al.(2024{\natexlab{b}})Dou, Liu, Zeng, Guo, Zhou, Mao, Jin, Lu, and Lin}]{dou-etal-2024-sailor}
Longxu Dou, Qian Liu, Guangtao Zeng, Jia Guo, Jiahui Zhou, Xin Mao, Ziqi Jin, Wei Lu, and Min Lin. 2024{\natexlab{b}}.
\newblock \href {https://doi.org/10.18653/v1/2024.emnlp-demo.45} {Sailor: Open language models for south-{E}ast {A}sia}.
\newblock In \emph{Proceedings of the 2024 Conference on Empirical Methods in Natural Language Processing: System Demonstrations}, pages 424--435, Miami, Florida, USA. Association for Computational Linguistics.

\bibitem[{Dou et~al.(2025)Dou, Liu, Zhou, Chen, Wang, Jin, Liu, Zhu, Du, Yang et~al.}]{dou2025sailor2}
Longxu Dou, Qian Liu, Fan Zhou, Changyu Chen, Zili Wang, Ziqi Jin, Zichen Liu, Tongyao Zhu, Cunxiao Du, Penghui Yang, and 1 others. 2025.
\newblock Sailor2: Sailing in south-east asia with inclusive multilingual llms.
\newblock \emph{arXiv preprint arXiv:2502.12982}.

\bibitem[{Fishman et~al.(2024)Fishman, Chmiel, Banner, and Soudry}]{fishman2024scaling}
Maxim Fishman, Brian Chmiel, Ron Banner, and Daniel Soudry. 2024.
\newblock Scaling fp8 training to trillion-token llms.
\newblock \emph{arXiv preprint arXiv:2409.12517}.

\bibitem[{Gao et~al.(2024)Gao, Tow, Abbasi, Biderman, Black, DiPofi, Foster, Golding, Hsu, Le~Noac'h, Li, McDonell, Muennighoff, Ociepa, Phang, Reynolds, Schoelkopf, Skowron, Sutawika, Tang, Thite, Wang, Wang, and Zou}]{eval-harness}
Leo Gao, Jonathan Tow, Baber Abbasi, Stella Biderman, Sid Black, Anthony DiPofi, Charles Foster, Laurence Golding, Jeffrey Hsu, Alain Le~Noac'h, Haonan Li, Kyle McDonell, Niklas Muennighoff, Chris Ociepa, Jason Phang, Laria Reynolds, Hailey Schoelkopf, Aviya Skowron, Lintang Sutawika, and 5 others. 2024.
\newblock \href {https://doi.org/10.5281/zenodo.12608602} {The language model evaluation harness}.

\bibitem[{Gelles et~al.(2024)Gelles, Kinoshita, Musser, and Dunham}]{gelles2024resource}
Rebecca Gelles, Veronica Kinoshita, Micah Musser, and James Dunham. 2024.
\newblock Resource democratization: is compute the binding constraint on ai research?
\newblock In \emph{Proceedings of the Thirty-Eighth AAAI Conference on Artificial Intelligence and Thirty-Sixth Conference on Innovative Applications of Artificial Intelligence and Fourteenth Symposium on Educational Advances in Artificial Intelligence}, pages 19840--19848.

\bibitem[{Gemma-Team et~al.(2024)Gemma-Team, Riviere, Pathak, Sessa, Hardin, Bhupatiraju, Hussenot, Mesnard, Shahriari, Ram{\'e} et~al.}]{team2024gemma}
Gemma-Team, Morgane Riviere, Shreya Pathak, Pier~Giuseppe Sessa, Cassidy Hardin, Surya Bhupatiraju, L{\'e}onard Hussenot, Thomas Mesnard, Bobak Shahriari, Alexandre Ram{\'e}, and 1 others. 2024.
\newblock Gemma 2: Improving open language models at a practical size.
\newblock \emph{arXiv preprint arXiv:2408.00118}.

\bibitem[{Glasze et~al.(2023)Glasze, Cattaruzza, Douzet, Dammann, Bertran, B{\^o}mont, Braun, Danet, Desforges, G{\'e}ry et~al.}]{glasze2023contested}
Georg Glasze, Ama{\"e}l Cattaruzza, Fr{\'e}d{\'e}rick Douzet, Finn Dammann, Marie-Gabrielle Bertran, Clotilde B{\^o}mont, Matthias Braun, Didier Danet, Alix Desforges, Aude G{\'e}ry, and 1 others. 2023.
\newblock Contested spatialities of digital sovereignty.
\newblock \emph{Geopolitics}, 28(2):919--958.

\bibitem[{Gosal et~al.(2024)Gosal, Xu, Ramakrishnan, Joshi, Sheinin, Mishra, Vassilieva, Hestness, Sengupta, Sahu et~al.}]{gosal2024bilingual}
Gurpreet Gosal, Yishi Xu, Gokul Ramakrishnan, Rituraj Joshi, Avraham Sheinin, Biswajit Mishra, Natalia Vassilieva, Joel Hestness, Neha Sengupta, Sunil~Kumar Sahu, and 1 others. 2024.
\newblock Bilingual adaptation of monolingual foundation models.
\newblock \emph{arXiv preprint arXiv:2407.12869}.

\bibitem[{Grattafiori et~al.(2024)Grattafiori, Dubey, Jauhri, Pandey, Kadian, Al-Dahle, Letman, Mathur, Schelten, Vaughan et~al.}]{grattafiori2024llama}
Aaron Grattafiori, Abhimanyu Dubey, Abhinav Jauhri, Abhinav Pandey, Abhishek Kadian, Ahmad Al-Dahle, Aiesha Letman, Akhil Mathur, Alan Schelten, Alex Vaughan, and 1 others. 2024.
\newblock The llama 3 herd of models.
\newblock \emph{arXiv preprint arXiv:2407.21783}.

\bibitem[{Ham et~al.(2020)Ham, Choe, Park, Choi, and Soh}]{ham-etal-2020-kornli}
Jiyeon Ham, Yo~Joong Choe, Kyubyong Park, Ilji Choi, and Hyungjoon Soh. 2020.
\newblock \href {https://doi.org/10.18653/v1/2020.findings-emnlp.39} {{K}or{NLI} and {K}or{STS}: New benchmark datasets for {K}orean natural language understanding}.
\newblock In \emph{Findings of the Association for Computational Linguistics: EMNLP 2020}, pages 422--430, Online. Association for Computational Linguistics.

\bibitem[{Heafield(2011)}]{heafield2011kenlm}
Kenneth Heafield. 2011.
\newblock Kenlm: Faster and smaller language model queries.
\newblock In \emph{Proceedings of the sixth workshop on statistical machine translation}, pages 187--197.

\bibitem[{Hendrycks et~al.(2021)Hendrycks, Burns, Basart, Zou, Mazeika, Song, and Steinhardt}]{hendrycks2021mmlu}
Dan Hendrycks, Collin Burns, Steven Basart, Andy Zou, Mantas Mazeika, Dawn Song, and Jacob Steinhardt. 2021.
\newblock \href {https://openreview.net/forum?id=d7KBjmI3GmQ} {Measuring massive multitask language understanding}.
\newblock In \emph{International Conference on Learning Representations}.

\bibitem[{Izsak et~al.(2021)Izsak, Berchansky, and Levy}]{izsak-etal-2021-train}
Peter Izsak, Moshe Berchansky, and Omer Levy. 2021.
\newblock \href {https://doi.org/10.18653/v1/2021.emnlp-main.831} {How to train {BERT} with an academic budget}.
\newblock In \emph{Proceedings of the 2021 Conference on Empirical Methods in Natural Language Processing}, pages 10644--10652, Online and Punta Cana, Dominican Republic. Association for Computational Linguistics.

\bibitem[{Jang et~al.(2022)Jang, Kim, Kwon, and Davis}]{jang-etal-2022-kobest}
Myeongjun Jang, Dohyung Kim, Deuk~Sin Kwon, and Eric Davis. 2022.
\newblock \href {https://aclanthology.org/2022.coling-1.325/} {{K}o{BEST}: {K}orean balanced evaluation of significant tasks}.
\newblock In \emph{Proceedings of the 29th International Conference on Computational Linguistics}, pages 3697--3708, Gyeongju, Republic of Korea. International Committee on Computational Linguistics.

\bibitem[{Jiang et~al.(2023)Jiang, Sablayrolles, Mensch, Bamford, Chaplot, de~las Casas, Bressand, Lengyel, Lample, Saulnier, Lavaud, Lachaux, Stock, Scao, Lavril, Wang, Lacroix, and Sayed}]{jiang2023mistral7b}
Albert~Q. Jiang, Alexandre Sablayrolles, Arthur Mensch, Chris Bamford, Devendra~Singh Chaplot, Diego de~las Casas, Florian Bressand, Gianna Lengyel, Guillaume Lample, Lucile Saulnier, Lélio~Renard Lavaud, Marie-Anne Lachaux, Pierre Stock, Teven~Le Scao, Thibaut Lavril, Thomas Wang, Timothée Lacroix, and William~El Sayed. 2023.
\newblock \href {https://arxiv.org/abs/2310.06825} {Mistral 7b}.

\bibitem[{Kang and Kim(2024)}]{kang2024krhumaneval}
Deokyeong Kang and Taeuk Kim. 2024.
\newblock Analysis of language models in korean program synthesis based on the kr-humaneval benchmark.
\newblock In \emph{Annual Conference on Human and Language Technology}, pages 245--250. Human and Language Technology.

\bibitem[{Kim et~al.(2024)Kim, Kim, Park, Lee, Song, Kim, Kim, Kim, Lee, Kim, Ahn, Yang, Lee, Park, Gim, Cha, Lee, and Kim}]{kim-etal-2024-solar}
Sanghoon Kim, Dahyun Kim, Chanjun Park, Wonsung Lee, Wonho Song, Yunsu Kim, Hyeonwoo Kim, Yungi Kim, Hyeonju Lee, Jihoo Kim, Changbae Ahn, Seonghoon Yang, Sukyung Lee, Hyunbyung Park, Gyoungjin Gim, Mikyoung Cha, Hwalsuk Lee, and Sunghun Kim. 2024.
\newblock \href {https://doi.org/10.18653/v1/2024.naacl-industry.3} {{SOLAR} 10.7{B}: Scaling large language models with simple yet effective depth up-scaling}.
\newblock In \emph{Proceedings of the 2024 Conference of the North American Chapter of the Association for Computational Linguistics: Human Language Technologies (Volume 6: Industry Track)}, pages 23--35, Mexico City, Mexico. Association for Computational Linguistics.

\bibitem[{Kiulian et~al.(2024)Kiulian, Polishko, Khandoga, Kostiuk, Gabrielli, Gaga{\l}a, Zaraket, Obaida, Garud, Mak et~al.}]{kiulian2024english}
Artur Kiulian, Anton Polishko, Mykola Khandoga, Yevhen Kostiuk, Guillermo Gabrielli, {\L}ukasz Gaga{\l}a, Fadi Zaraket, Qusai~Abu Obaida, Hrishikesh Garud, Wendy Wing~Yee Mak, and 1 others. 2024.
\newblock From english-centric to effective bilingual: Llms with custom tokenizers for underrepresented languages.
\newblock \emph{arXiv preprint arXiv:2410.18836}.

\bibitem[{Ko et~al.(2023)Ko, Yang, Ryu, Choi, Yang, Hyun, Park, and Park}]{ko2023technical}
Hyunwoong Ko, Kichang Yang, Minho Ryu, Taekyoon Choi, Seungmu Yang, Jiwung Hyun, Sungho Park, and Kyubyong Park. 2023.
\newblock A technical report for polyglot-ko: Open-source large-scale korean language models.
\newblock \emph{arXiv preprint arXiv:2306.02254}.

\bibitem[{Kwon et~al.(2023)Kwon, Li, Zhuang, Sheng, Zheng, Yu, Gonzalez, Zhang, and Stoica}]{kwon2023efficient}
Woosuk Kwon, Zhuohan Li, Siyuan Zhuang, Ying Sheng, Lianmin Zheng, Cody~Hao Yu, Joseph~E. Gonzalez, Hao Zhang, and Ion Stoica. 2023.
\newblock Efficient memory management for large language model serving with pagedattention.
\newblock In \emph{Proceedings of the ACM SIGOPS 29th Symposium on Operating Systems Principles}.

\bibitem[{Lee et~al.(2025)Lee, Kim, Park, and Lee}]{lee2025dna10technicalreport}
Jungyup Lee, Jemin Kim, Sang Park, and SeungJae Lee. 2025.
\newblock \href {https://arxiv.org/abs/2501.10648} {Dna 1.0 technical report}.
\newblock \emph{Preprint}, arXiv:2501.10648.

\bibitem[{LG-Research et~al.(2024)LG-Research, An, Bae, Choi, Choi, Choi, Hong, Hwang, Jeon, Jo et~al.}]{research2024exaone}
LG-Research, Soyoung An, Kyunghoon Bae, Eunbi Choi, Kibong Choi, Stanley~Jungkyu Choi, Seokhee Hong, Junwon Hwang, Hyojin Jeon, Gerrard~Jeongwon Jo, and 1 others. 2024.
\newblock Exaone 3.5: Series of large language models for real-world use cases.
\newblock \emph{arXiv preprint arXiv:2412.04862}.

\bibitem[{Li et~al.(2024)Li, Fang, Smyrnis, Ivgi, Jordan, Gadre, Bansal, Guha, Keh, Arora et~al.}]{li2024datacomp}
Jeffrey Li, Alex Fang, Georgios Smyrnis, Maor Ivgi, Matt Jordan, Samir~Yitzhak Gadre, Hritik Bansal, Etash Guha, Sedrick~Scott Keh, Kushal Arora, and 1 others. 2024.
\newblock Datacomp-lm: In search of the next generation of training sets for language models.
\newblock \emph{Advances in Neural Information Processing Systems}, 37:14200--14282.

\bibitem[{LI et~al.(2024)LI, Beeching, Tunstall, Lipkin, Soletskyi, Huang, Rasul, Yu, Jiang, Shen, Qin, Dong, Zhou, Fleureau, Lample, and Polu}]{numina_math_datasets}
Jia LI, Edward Beeching, Lewis Tunstall, Ben Lipkin, Roman Soletskyi, Shengyi~Costa Huang, Kashif Rasul, Longhui Yu, Albert Jiang, Ziju Shen, Zihan Qin, Bin Dong, Li~Zhou, Yann Fleureau, Guillaume Lample, and Stanislas Polu. 2024.
\newblock Numinamath.
\newblock \url{[https://huggingface.co/AI-MO/NuminaMath-CoT](https://github.com/project-numina/aimo-progress-prize/blob/main/report/numina_dataset.pdf)}.

\bibitem[{Lian et~al.(2023{\natexlab{a}})Lian, Goodson, Pentland, Cook, Vong, and "Teknium"}]{OpenOrca}
Wing Lian, Bleys Goodson, Eugene Pentland, Austin Cook, Chanvichet Vong, and "Teknium". 2023{\natexlab{a}}.
\newblock Openorca: An open dataset of gpt augmented flan reasoning traces.
\newblock \url{https://https://huggingface.co/datasets/Open-Orca/OpenOrca}.

\bibitem[{Lian et~al.(2023{\natexlab{b}})Lian, Wang, Goodson, Pentland, Cook, Vong, and "Teknium"}]{SlimOrca}
Wing Lian, Guan Wang, Bleys Goodson, Eugene Pentland, Austin Cook, Chanvichet Vong, and "Teknium". 2023{\natexlab{b}}.
\newblock \href {https://https://huggingface.co/Open-Orca/SlimOrca} {Slimorca: An open dataset of gpt-4 augmented flan reasoning traces, with verification}.

\bibitem[{Lim et~al.(2022)Lim, Cho, Hur, Yim, Ko, Chun, Choi, Jeong, Yu, Shin, Jang, Kim, and Lee}]{MWP_KR_DATA}
Soyoung Lim, Heecheol Cho, Taeil Hur, Jiyeon Yim, Taeyoung Ko, Tae-Hyun Chun, Eunjin Choi, Jiyoung Jeong, Yonggyun Yu, Donghyun Shin, GyeongHwan Jang, Minjong Kim, and Sangwon Lee. 2022.
\newblock Mwp\_kr\_data, dataset for math word problems in korean language.
\newblock \url{https://github.com/jkc-ai/mwp_kr_data}.

\bibitem[{Longpre et~al.(2023)Longpre, Hou, Vu, Webson, Chung, Tay, Zhou, Le, Zoph, Wei, and Roberts}]{longpre2023flan}
Shayne Longpre, Le~Hou, Tu~Vu, Albert Webson, Hyung~Won Chung, Yi~Tay, Denny Zhou, Quoc~V. Le, Barret Zoph, Jason Wei, and Adam Roberts. 2023.
\newblock \href {https://arxiv.org/abs/2301.13688} {The flan collection: Designing data and methods for effective instruction tuning}.
\newblock \emph{Preprint}, arXiv:2301.13688.

\bibitem[{Meng et~al.(2024)Meng, Xia, and Chen}]{meng2024simposimplepreferenceoptimization}
Yu~Meng, Mengzhou Xia, and Danqi Chen. 2024.
\newblock \href {https://proceedings.neurips.cc/paper_files/paper/2024/file/e099c1c9699814af0be873a175361713-Paper-Conference.pdf} {Simpo: Simple preference optimization with a reference-free reward}.
\newblock In \emph{Advances in Neural Information Processing Systems}, volume~37, pages 124198--124235. Curran Associates, Inc.

\bibitem[{Mihaylov et~al.(2018)Mihaylov, Clark, Khot, and Sabharwal}]{mihaylov-etal-2018-suit}
Todor Mihaylov, Peter Clark, Tushar Khot, and Ashish Sabharwal. 2018.
\newblock \href {https://doi.org/10.18653/v1/D18-1260} {Can a suit of armor conduct electricity? a new dataset for open book question answering}.
\newblock In \emph{Proceedings of the 2018 Conference on Empirical Methods in Natural Language Processing}, pages 2381--2391, Brussels, Belgium. Association for Computational Linguistics.

\bibitem[{Minixhofer et~al.(2021)Minixhofer, Paischer, and Rekabsaz}]{minixhofer2021wechsel}
Benjamin Minixhofer, Fabian Paischer, and Navid Rekabsaz. 2021.
\newblock Wechsel: Effective initialization of subword embeddings for cross-lingual transfer of monolingual language models.
\newblock \emph{arXiv preprint arXiv:2112.06598}.

\bibitem[{Mitra et~al.(2024)Mitra, Khanpour, Rosset, and Awadallah}]{mitra2024orca}
Arindam Mitra, Hamed Khanpour, Corby Rosset, and Ahmed Awadallah. 2024.
\newblock Orca-math: Unlocking the potential of slms in grade school math.
\newblock \emph{arXiv preprint arXiv:2402.14830}.

\bibitem[{Mukherjee et~al.(2023)Mukherjee, Mitra, Jawahar, Agarwal, Palangi, and Awadallah}]{mukherjee2023orca}
Subhabrata Mukherjee, Arindam Mitra, Ganesh Jawahar, Sahaj Agarwal, Hamid Palangi, and Ahmed Awadallah. 2023.
\newblock \href {https://arxiv.org/abs/2306.02707} {Orca: Progressive learning from complex explanation traces of gpt-4}.
\newblock \emph{Preprint}, arXiv:2306.02707.

\bibitem[{NVIDIA()}]{nvidiah100}
NVIDIA.
\newblock Nvidia h100 tensor core gpu architecture.
\newblock \url{https://resources.nvidia.com/en-us-data-center-overview/gtc22-whitepaper-hopper}.

\bibitem[{NVIDIA(2025)}]{nvidia_transformer_engine}
NVIDIA. 2025.
\newblock Transformer engine.
\newblock \url{https://github.com/NVIDIA/TransformerEngine}.
\newblock Accessed: 2025-05-12.

\bibitem[{Ouyang et~al.(2022)Ouyang, Wu, Jiang, Almeida, Wainwright, Mishkin, Zhang, Agarwal, Slama, Ray et~al.}]{ouyang2022training}
Long Ouyang, Jeffrey Wu, Xu~Jiang, Diogo Almeida, Carroll Wainwright, Pamela Mishkin, Chong Zhang, Sandhini Agarwal, Katarina Slama, Alex Ray, and 1 others. 2022.
\newblock Training language models to follow instructions with human feedback.
\newblock \emph{Advances in neural information processing systems}, 35:27730--27744.

\bibitem[{Paperno et~al.(2016)Paperno, Kruszewski, Lazaridou, Pham, Bernardi, Pezzelle, Baroni, Boleda, and Fern{\'a}ndez}]{paperno-etal-2016-lambada}
Denis Paperno, Germ{\'a}n Kruszewski, Angeliki Lazaridou, Ngoc~Quan Pham, Raffaella Bernardi, Sandro Pezzelle, Marco Baroni, Gemma Boleda, and Raquel Fern{\'a}ndez. 2016.
\newblock \href {https://doi.org/10.18653/v1/P16-1144} {The {LAMBADA} dataset: Word prediction requiring a broad discourse context}.
\newblock In \emph{Proceedings of the 54th Annual Meeting of the Association for Computational Linguistics (Volume 1: Long Papers)}, pages 1525--1534, Berlin, Germany. Association for Computational Linguistics.

\bibitem[{Park et~al.(2021)Park, Moon, Kim, Cho, Han, Park, Song, Kim, Song, Oh et~al.}]{park2021klue}
Sungjoon Park, Jihyung Moon, Sungdong Kim, Won~Ik Cho, Jiyoon Han, Jangwon Park, Chisung Song, Junseong Kim, Yongsook Song, Taehwan Oh, and 1 others. 2021.
\newblock Klue: Korean language understanding evaluation.
\newblock \emph{arXiv preprint arXiv:2105.09680}.

\bibitem[{Paszke(2019)}]{paszke2019pytorch}
A~Paszke. 2019.
\newblock Pytorch: An imperative style, high-performance deep learning library.
\newblock \emph{arXiv preprint arXiv:1912.01703}.

\bibitem[{Peng et~al.(2023)Peng, Wu, Wei, Zhao, Yang, Liu, Xiong, Yang, Ni, Hu et~al.}]{peng2023fp8}
Houwen Peng, Kan Wu, Yixuan Wei, Guoshuai Zhao, Yuxiang Yang, Ze~Liu, Yifan Xiong, Ziyue Yang, Bolin Ni, Jingcheng Hu, and 1 others. 2023.
\newblock Fp8-lm: Training fp8 large language models.
\newblock \emph{arXiv preprint arXiv:2310.18313}.

\bibitem[{Qwen et~al.(2025)Qwen, :, Yang, Yang, Zhang, Hui, Zheng, Yu, Li, Liu, Huang, Wei, Lin, Yang, Tu, Zhang, Yang, Yang, Zhou, Lin, Dang, Lu, Bao, Yang, Yu, Li, Xue, Zhang, Zhu, Men, Lin, Li, Tang, Xia, Ren, Ren, Fan, Su, Zhang, Wan, Liu, Cui, Zhang, and Qiu}]{qwen25technicalreport}
Qwen, :, An~Yang, Baosong Yang, Beichen Zhang, Binyuan Hui, Bo~Zheng, Bowen Yu, Chengyuan Li, Dayiheng Liu, Fei Huang, Haoran Wei, Huan Lin, Jian Yang, Jianhong Tu, Jianwei Zhang, Jianxin Yang, Jiaxi Yang, Jingren Zhou, and 25 others. 2025.
\newblock \href {https://arxiv.org/abs/2412.15115} {Qwen2.5 technical report}.
\newblock \emph{Preprint}, arXiv:2412.15115.

\bibitem[{Rae et~al.(2021)Rae, Borgeaud, Cai, Millican, Hoffmann, Song, Aslanides, Henderson, Ring, Young et~al.}]{rae2021scaling}
Jack~W Rae, Sebastian Borgeaud, Trevor Cai, Katie Millican, Jordan Hoffmann, Francis Song, John Aslanides, Sarah Henderson, Roman Ring, Susannah Young, and 1 others. 2021.
\newblock Scaling language models: Methods, analysis \& insights from training gopher.
\newblock \emph{arXiv preprint arXiv:2112.11446}.

\bibitem[{Rafailov et~al.(2023)Rafailov, Sharma, Mitchell, Manning, Ermon, and Finn}]{rafailov2023direct}
Rafael Rafailov, Archit Sharma, Eric Mitchell, Christopher~D Manning, Stefano Ermon, and Chelsea Finn. 2023.
\newblock Direct preference optimization: Your language model is secretly a reward model.
\newblock \emph{Advances in Neural Information Processing Systems}, 36:53728--53741.

\bibitem[{Rajbhandari et~al.(2020)Rajbhandari, Rasley, Ruwase, and He}]{rajbhandari2020zero}
Samyam Rajbhandari, Jeff Rasley, Olatunji Ruwase, and Yuxiong He. 2020.
\newblock Zero: Memory optimizations toward training trillion parameter models.
\newblock In \emph{SC20: International Conference for High Performance Computing, Networking, Storage and Analysis}, pages 1--16. IEEE.

\bibitem[{Richardson(2007)}]{richardson2007beautiful}
Leonard Richardson. 2007.
\newblock \href {https://www.crummy.com/software/BeautifulSoup/bs4/doc/} {Beautiful soup documentation}.
\newblock \emph{April}.

\bibitem[{Roberts et~al.(2023)Roberts, Hine, and Floridi}]{roberts2023digital}
Huw Roberts, Emmie Hine, and Luciano Floridi. 2023.
\newblock Digital sovereignty, digital expansionism, and the prospects for global ai governance.
\newblock In \emph{Quo Vadis, Sovereignty? New Conceptual and Regulatory Boundaries in the Age of Digital China}, pages 51--75. Springer.

\bibitem[{Sakaguchi et~al.(2021)Sakaguchi, Bras, Bhagavatula, and Choi}]{sakaguchi2021winogrande}
Keisuke Sakaguchi, Ronan~Le Bras, Chandra Bhagavatula, and Yejin Choi. 2021.
\newblock Winogrande: An adversarial winograd schema challenge at scale.
\newblock \emph{Communications of the ACM}, 64(9):99--106.

\bibitem[{Saura~Garc{\'\i}a(2024{\natexlab{a}})}]{saura2024datafeudalism}
Carlos Saura~Garc{\'\i}a. 2024{\natexlab{a}}.
\newblock Datafeudalism: the domination of modern societies by big tech companies.
\newblock \emph{Philosophy \& Technology}, 37(3):90.

\bibitem[{Saura~Garc{\'\i}a(2024{\natexlab{b}})}]{saura2024digital}
Carlos Saura~Garc{\'\i}a. 2024{\natexlab{b}}.
\newblock Digital expansionism and big tech companies: consequences in democracies of the european union.
\newblock \emph{Humanities and Social Sciences Communications}, 11(1):1--8.

\bibitem[{Soldaini et~al.(2024)Soldaini, Kinney, Bhagia, Schwenk, Atkinson, Authur, Bogin, Chandu, Dumas, Elazar et~al.}]{soldaini2024dolma}
Luca Soldaini, Rodney Kinney, Akshita Bhagia, Dustin Schwenk, David Atkinson, Russell Authur, Ben Bogin, Khyathi Chandu, Jennifer Dumas, Yanai Elazar, and 1 others. 2024.
\newblock Dolma: An open corpus of three trillion tokens for language model pretraining research.
\newblock \emph{arXiv preprint arXiv:2402.00159}.

\bibitem[{Son et~al.(2024)Son, Lee, Kim, Kim, Muennighoff, Choi, Park, Yoo, and Biderman}]{son2024kmmlu}
Guijin Son, Hanwool Lee, Sungdong Kim, Seungone Kim, Niklas Muennighoff, Taekyoon Choi, Cheonbok Park, Kang~Min Yoo, and Stella Biderman. 2024.
\newblock Kmmlu: Measuring massive multitask language understanding in korean.
\newblock \emph{arXiv preprint arXiv:2402.11548}.

\bibitem[{Son et~al.(2025{\natexlab{a}})Son, Lee, Kim, Kim, Muennighoff, Choi, Park, Yoo, and Biderman}]{son-etal-2025-kmmlu}
Guijin Son, Hanwool Lee, Sungdong Kim, Seungone Kim, Niklas Muennighoff, Taekyoon Choi, Cheonbok Park, Kang~Min Yoo, and Stella Biderman. 2025{\natexlab{a}}.
\newblock \href {https://aclanthology.org/2025.naacl-long.206/} {{KMMLU}: Measuring massive multitask language understanding in {K}orean}.
\newblock In \emph{Proceedings of the 2025 Conference of the Nations of the Americas Chapter of the Association for Computational Linguistics: Human Language Technologies (Volume 1: Long Papers)}, pages 4076--4104, Albuquerque, New Mexico. Association for Computational Linguistics.

\bibitem[{Son et~al.(2025{\natexlab{b}})Son, Kim, and Lee}]{son2025fed}
Youngjun Son, Chaewon Kim, and Jaejin Lee. 2025{\natexlab{b}}.
\newblock Fed: Fast and efficient dataset deduplication framework with gpu acceleration.
\newblock \emph{arXiv preprint arXiv:2501.01046}.

\bibitem[{Taori et~al.(2023)Taori, Gulrajani, Zhang, Dubois, Li, Guestrin, Liang, and Hashimoto}]{alpaca}
Rohan Taori, Ishaan Gulrajani, Tianyi Zhang, Yann Dubois, Xuechen Li, Carlos Guestrin, Percy Liang, and Tatsunori~B. Hashimoto. 2023.
\newblock Stanford alpaca: An instruction-following llama model.
\newblock \url{https://github.com/tatsu-lab/stanford_alpaca}.

\bibitem[{Von~Werra et~al.(2022)Von~Werra, Tunstall, Thakur, Luccioni, Thrush, Piktus, Marty, Rajani, Mustar, and Ngo}]{von-werra-etal-2022-evaluate}
Leandro Von~Werra, Lewis Tunstall, Abhishek Thakur, Sasha Luccioni, Tristan Thrush, Aleksandra Piktus, Felix Marty, Nazneen Rajani, Victor Mustar, and Helen Ngo. 2022.
\newblock \href {https://doi.org/10.18653/v1/2022.emnlp-demos.13} {Evaluate {\&} evaluation on the hub: Better best practices for data and model measurements}.
\newblock In \emph{Proceedings of the 2022 Conference on Empirical Methods in Natural Language Processing: System Demonstrations}, pages 128--136, Abu Dhabi, UAE. Association for Computational Linguistics.

\bibitem[{Weber et~al.(2024)Weber, Fu, Anthony, Oren, Adams, Alexandrov, Lyu, Nguyen, Yao, Adams et~al.}]{weber2024redpajama}
Maurice Weber, Dan Fu, Quentin Anthony, Yonatan Oren, Shane Adams, Anton Alexandrov, Xiaozhong Lyu, Huu Nguyen, Xiaozhe Yao, Virginia Adams, and 1 others. 2024.
\newblock Redpajama: an open dataset for training large language models.
\newblock \emph{Advances in neural information processing systems}, 37:116462--116492.

\bibitem[{Wei et~al.(2022{\natexlab{a}})Wei, Bosma, Zhao, Guu, Yu, Lester, Du, Dai, and Le}]{wei2022finetunedlanguagemodelszeroshot}
Jason Wei, Maarten Bosma, Vincent~Y. Zhao, Kelvin Guu, Adams~Wei Yu, Brian Lester, Nan Du, Andrew~M. Dai, and Quoc~V. Le. 2022{\natexlab{a}}.
\newblock \href {https://openreview.net/forum?id=gEZrGCozdqR} {Finetuned language models are zero-shot learners}.
\newblock In \emph{The Tenth International Conference on Learning Representations, {ICLR} 2022, Virtual Event, April 25-29, 2022}.

\bibitem[{Wei et~al.(2022{\natexlab{b}})Wei, Wang, Schuurmans, Bosma, Xia, Chi, Le, Zhou et~al.}]{wei2022chain}
Jason Wei, Xuezhi Wang, Dale Schuurmans, Maarten Bosma, Fei Xia, Ed~Chi, Quoc~V Le, Denny Zhou, and 1 others. 2022{\natexlab{b}}.
\newblock Chain-of-thought prompting elicits reasoning in large language models.
\newblock \emph{Advances in neural information processing systems}, 35:24824--24837.

\bibitem[{Wikimedia()}]{wikidump}
Wikimedia.
\newblock \href {https://dumps.wikimedia.org} {Wikimedia downloads}.

\bibitem[{Xi et~al.(2024)Xi, Wu, Fan, Chen, Gu, Yu, Qu, Liu, Jiang, Chen et~al.}]{xi2024practice}
Ningyuan Xi, Yetao Wu, Kun Fan, Teng Chen, Qingqing Gu, Peng Yu, Jinxian Qu, Chenxi Liu, Zhonglin Jiang, Yong Chen, and 1 others. 2024.
\newblock A practice of post-training on llama-3 70b with optimal selection of additional language mixture ratio.
\newblock \emph{arXiv preprint arXiv:2409.06624}.

\bibitem[{Xia et~al.(2025)Xia, Shen, Wang, Liu, Sun, Wu, Hu, and Xu}]{xia2025leetcodedataset}
Yunhui Xia, Wei Shen, Yan Wang, Jason~Klein Liu, Huifeng Sun, Siyue Wu, Jian Hu, and Xiaolong Xu. 2025.
\newblock Leetcodedataset: A temporal dataset for robust evaluation and efficient training of code llms.
\newblock \emph{arXiv preprint arXiv:2504.14655}.

\bibitem[{Yoo et~al.(2024)Yoo, Han, In, Jeon, Jeong, Kang, Kim, Kim, Kim, Kim et~al.}]{yoo2024hyperclova}
Kang~Min Yoo, Jaegeun Han, Sookyo In, Heewon Jeon, Jisu Jeong, Jaewook Kang, Hyunwook Kim, Kyung-Min Kim, Munhyong Kim, Sungju Kim, and 1 others. 2024.
\newblock Hyperclova x technical report.
\newblock \emph{arXiv preprint arXiv:2404.01954}.

\bibitem[{Zellers et~al.(2019)Zellers, Holtzman, Bisk, Farhadi, and Choi}]{zellers2019hellaswag}
Rowan Zellers, Ari Holtzman, Yonatan Bisk, Ali Farhadi, and Yejin Choi. 2019.
\newblock Hellaswag: Can a machine really finish your sentence?
\newblock \emph{arXiv preprint arXiv:1905.07830}.

\bibitem[{Zhou et~al.(2023)Zhou, Lu, Mishra, Brahma, Basu, Luan, Zhou, and Hou}]{zhou2023instruction}
Jeffrey Zhou, Tianjian Lu, Swaroop Mishra, Siddhartha Brahma, Sujoy Basu, Yi~Luan, Denny Zhou, and Le~Hou. 2023.
\newblock Instruction-following evaluation for large language models.
\newblock \emph{arXiv preprint arXiv:2311.07911}.

\end{thebibliography}
